%% file: main.tex
\documentclass[10pt,twocolumn,letterpaper]{article}

\usepackage[pagenumbers]{cvpr} % To force page numbers, e.g. for an arXiv version

\input{preamble}

\definecolor{cvprblue}{rgb}{0.21,0.49,0.74}
\usepackage[pagebackref,breaklinks,colorlinks,allcolors=cvprblue]{hyperref}

\def\confName{CVPR}
\def\confYear{2027}

\title{Compact SSM Hidden States for Accurate and Efficient Event-Based Vision}

\author{%
  Nikola Zubić\textsuperscript{1} \quad
  Haiqing Hao\textsuperscript{2} \quad
  Davide Scaramuzza\textsuperscript{1}\\[0.3em]
  {\small \textsuperscript{1}\,Robotics and Perception Group, University of Zurich}\\
  {\small \textsuperscript{2}\,Department of Precision Instrument, Tsinghua University}%
}

\begin{document}
\maketitle
% Queue a normal top float for the second (right) column of page one.
% Do not use [!t]: it overrides the first-column top-float suppression.
\suppressfloats[t]
\begin{figure}[t]
  \centering
  \resizebox{\columnwidth}{!}{\input{figures/teaser}}
    \caption{\textbf{Best of both worlds: rich spatial context at low energy cost.}
  \textbf{(a)}~Dense packet methods collect events into an image-like tensor
  and apply channel mixing at every spatial location. This keeps rich spatial
  context and thus high accuracy, but repeating the mixing at every location
  makes the energy cost high.
  \textbf{(b)}~Event-by-event methods update only a small part of the network
  when each event arrives. This keeps the energy cost per event low, but
  accuracy is lower.
  \textbf{(c)}~Our detector processes event packets, but it first compresses
  the many spatial tokens of a packet ($N$) into a much smaller set of compact
  SSM hidden states ($S$). The mixer's main channel mixing and gating run only
  on these few states. The result is then expanded back to all $N$ locations,
  so the detector can still locate objects. We therefore get the best of both
  worlds: the rich spatial context of packets, and thus high accuracy, at a
  low energy cost.}
  \label{fig:teaser}
\end{figure}
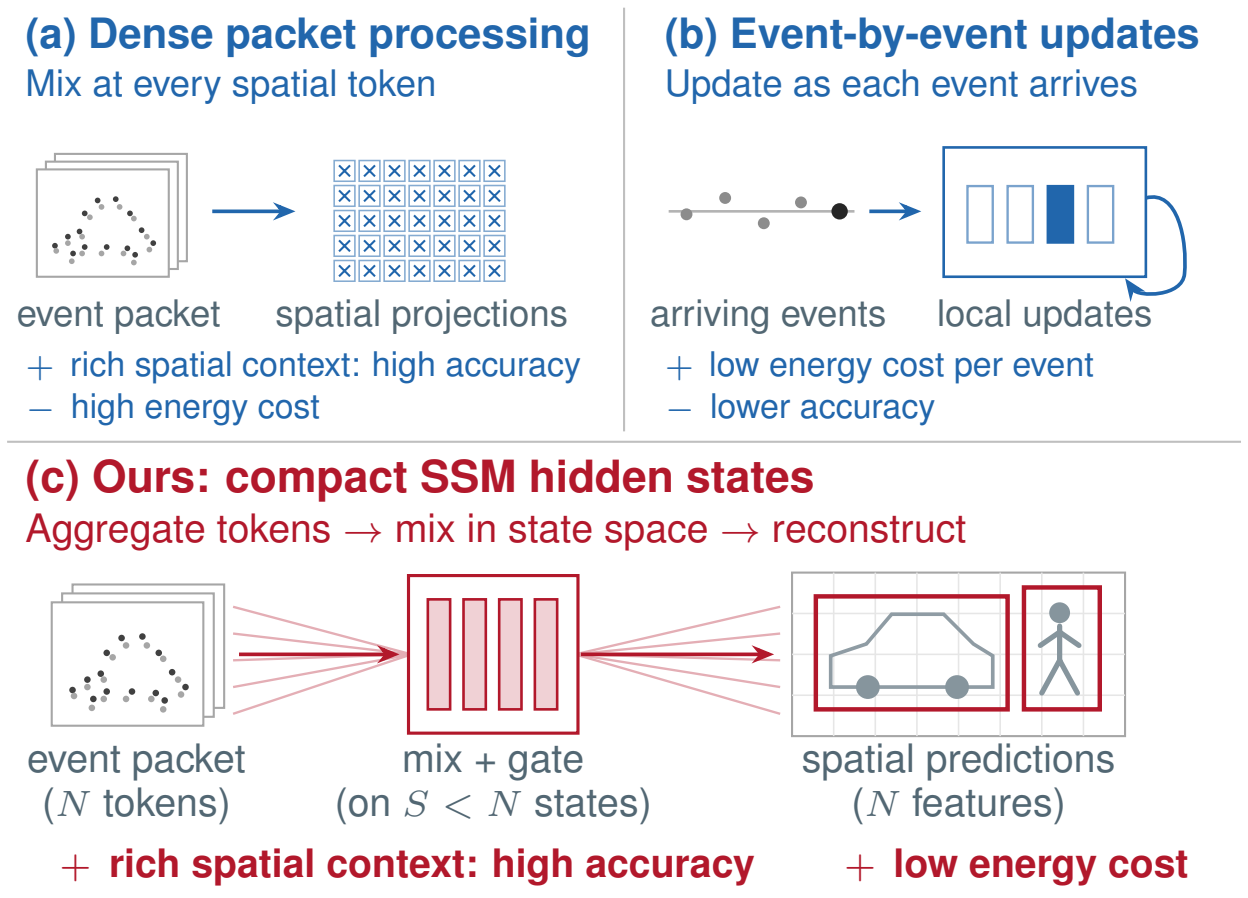
\input{sec/0_abstract}
\input{sec/1_intro}
\input{sec/2_related}
\input{sec/3_method}
% Completed experiments and their generated comparison tables.
\input{sec/2_experiments}

\input{sec/5_conclusion}
\input{sec/6_LLM_usage}
\input{sec/X_suppl}

% Keep the eight-page main paper separate from the references.
%\clearpage
{\small
\bibliographystyle{ieeenat_fullname}
\bibliography{main}
}


\end{document}

%% file: preamble.tex
\usepackage{microtype}
\usepackage{needspace}
\usepackage{etoolbox}
\input{figures/teaser_style}
\usepackage{enumitem}
\setlist[itemize]{leftmargin=*,itemsep=2pt,topsep=3pt}

%% file: figures/teaser_style.tex
\usepackage{tikz}
\usetikzlibrary{arrows.meta,calc}
\definecolor{TeaserInk}{HTML}{142D39}
\definecolor{TeaserMuted}{HTML}{526773}
\definecolor{TeaserCoral}{HTML}{C9684C}
\definecolor{TeaserBlue}{HTML}{557CB6}
\definecolor{TeaserTeal}{HTML}{007E78}
\definecolor{TeaserLight}{HTML}{EAF6F2}
\definecolor{TeaserRule}{HTML}{D4E3E5}

%% file: figures/teaser.tex
% Figure 1 (teaser). Load figures/teaser_style.tex before this file.
% Event dots, token grids and detections are illustrations, not dataset samples.
% Only the mixer's main channel projections/gating move into compact states.
% Prior approaches (a,b) are blue; ours (c) is red.
\definecolor{PriorColor}{HTML}{2166AC}  % blue (prior approaches)
\definecolor{OursColor}{HTML}{B2182B}   % red (ours)
\begin{tikzpicture}[
  x=1cm,y=1cm,
  font=\sffamily\fontsize{7.5}{8.5}\selectfont,
  text=TeaserInk,
  flow/.style={-{Stealth[length=1.5mm,width=1.2mm]},line width=.7pt},
  % Styles for panel (c)
  lbl/.style={font=\sffamily\fontsize{7.5}{8.5}\selectfont,text=TeaserMuted,align=center},
  ptitle/.style={font=\sffamily\bfseries\fontsize{8.5}{9.5}\selectfont,anchor=west},
  psub/.style={font=\sffamily\fontsize{7.5}{8.5}\selectfont,anchor=west},
  % Smaller styles for the half-width panels (a) and (b)
  lblS/.style={font=\sffamily\fontsize{7}{8}\selectfont,text=TeaserMuted,align=center},
  ptitleS/.style={font=\sffamily\bfseries\fontsize{8}{9}\selectfont,anchor=west},
  psubS/.style={font=\sffamily\fontsize{7}{8}\selectfont,anchor=west},
  proS/.style={font=\sffamily\fontsize{6.5}{7.5}\selectfont,anchor=west},
  sep/.style={draw=black!25,line width=.4pt},
  pics/packet/.style={code={%
    \foreach \s in {2,1,0}{
      \draw[fill=white,draw=black!35,line width=.4pt]
        (.07*\s,.06*\s) rectangle ++(1.02,.83);
    }
    \foreach \x/\y in {.14/.24,.24/.31,.33/.35,.37/.49,.47/.59,.61/.60,.72/.49,.81/.36,.90/.26,.70/.22,.54/.23,.37/.23,.28/.18,.76/.17}{
      \fill[black!75] (\x,\y) circle (.022);
    }
    \foreach \x/\y in {.15/.20,.24/.26,.35/.31,.41/.46,.50/.54,.64/.54,.75/.45,.85/.31,.87/.21,.67/.18,.50/.18,.35/.18,.27/.12,.75/.11}{
      \fill[black!35] (\x,\y) circle (.022);
    }
  }},
  pics/densegrid/.style={code={%
    \foreach \i in {0,...,6}{
      \foreach \j in {0,...,4}{
        \draw[fill=white,draw=PriorColor!55,line width=.3pt]
          (.205*\i,.205*\j) rectangle ++(.17,.17);
        \draw[PriorColor,line width=.4pt]
          (.205*\i+.04,.205*\j+.04) -- ++(.09,.09)
          (.205*\i+.04,.205*\j+.13) -- ++(.09,-.09);
      }
    }
  }}
]
  \path[use as bounding box] (0,.02) rectangle (8.30,6.03);

  % ---------------- (a) Dense packet processing ----------------
  \node[ptitleS,text=PriorColor] at (0,5.80) {(a) Dense packet processing};
  \node[psubS,text=PriorColor] at (0,5.47) {Mix at every spatial token};
  \pic[scale=.87] at (.20,4.19) {packet};
  \draw[flow,PriorColor] (1.38,4.63) -- (1.94,4.63);
  \pic[scale=.82] at (2.20,4.16) {densegrid};
  \node[lblS] at (.75,3.92) {event packet};
  \node[lblS] at (2.79,3.92) {spatial projections};
  \node[proS,text=PriorColor] at (0,3.58) {\(+\)\, rich spatial context: high accuracy};
  \node[proS,text=PriorColor] at (0,3.30) {\(-\)\, high energy cost}; % (a) con
  
  \draw[sep] (4.15,3.15) -- (4.15,6.00);

  % ---------------- (b) Event-by-event updates ----------------
  \node[ptitleS,text=PriorColor] at (4.30,5.80) {(b) Event-by-event updates};
  \node[psubS,text=PriorColor] at (4.30,5.47) {Update as each event arrives};
  \draw[black!30,line width=.5pt] (4.45,4.63) -- (5.70,4.63);
  \foreach \x/\y in {4.57/4.61,4.83/4.72,5.09/4.55,5.34/4.69}{
    \fill[black!45] (\x,\y) circle (.04);
  }
  \fill[black!85] (5.60,4.63) circle (.055);
  \draw[flow,PriorColor] (5.80,4.63) -- (6.15,4.63);
  \draw[fill=white,draw=PriorColor,line width=.5pt]
    (6.30,4.19) rectangle (7.66,5.06);
  \foreach \x in {6.46,6.73,7.27}{
    \draw[draw=PriorColor!60,line width=.45pt] (\x,4.42) rectangle ++(.17,.38);
  }
  \filldraw[fill=PriorColor,draw=PriorColor,line width=.45pt]
    (7.00,4.42) rectangle ++(.17,.38);
  \draw[flow,PriorColor] (7.66,4.72) to[out=15,in=85] (7.92,4.21)
    to[out=-95,in=-55] (7.52,4.18);
  \node[lblS] at (5.12,3.92) {arriving events};
  \node[lblS] at (6.98,3.92) {local updates};
  \node[proS,text=PriorColor] at (4.30,3.58) {\(+\)\, low energy cost per event};
  \node[proS,text=PriorColor] at (4.30,3.30) {\(-\)\, lower accuracy};

  \draw[sep] (0,3.08) -- (8.30,3.08);

  % ---------------- (c) Ours ----------------
  \node[ptitle,text=OursColor] at (0,2.82) {(c) Ours: compact SSM hidden states};
  \node[psub,text=OursColor] at (0,2.47)
    {Aggregate tokens \(\rightarrow\) mix in state space \(\rightarrow\) reconstruct};
  \pic at (.30,1.17) {packet};
  \foreach \y in {1.25,1.43,1.61,1.79,1.97}{
    \draw[OursColor!35,line width=.45pt] (1.52,\y) -- (2.72,1.65);
    \draw[OursColor!35,line width=.45pt] (3.82,1.65) -- (5.20,\y);
  }
  \draw[flow,OursColor] (1.56,1.65) -- (2.64,1.65);
  \draw[flow,OursColor] (3.86,1.65) -- (5.16,1.65);
  % S compact hidden states
  \draw[fill=white,draw=OursColor,line width=.6pt] (2.70,1.12) rectangle (3.84,2.18);
  \foreach \x in {2.83,3.07,3.31,3.55}{
    \filldraw[fill=OursColor!20,draw=OursColor,line width=.5pt]
      (\x,1.28) rectangle ++(.15,.74);
  }
  % Reconstructed spatial grid with illustrative detections
  \draw[fill=white,draw=black!35,line width=.4pt] (5.28,1.10) rectangle (7.52,2.20);
  \foreach \x in {5.56,5.84,6.12,6.40,6.68,6.96,7.24}{
    \draw[black!10,line width=.3pt] (\x,1.10) -- (\x,2.20);
  }
  \foreach \y in {1.375,1.65,1.925}{
    \draw[black!10,line width=.3pt] (5.28,\y) -- (7.52,\y);
  }
  \draw[TeaserMuted!65,line width=.7pt]
    (5.54,1.43) -- (5.54,1.64) -- (5.78,1.72) -- (5.95,1.92)
    -- (6.30,1.92) -- (6.50,1.72) -- (6.63,1.64) -- (6.63,1.43) -- cycle;
  \fill[TeaserMuted!70] (5.79,1.43) circle (.08);
  \fill[TeaserMuted!70] (6.39,1.43) circle (.08);
  \fill[TeaserMuted!70] (7.08,1.93) circle (.07);
  \draw[TeaserMuted!70,line width=1pt,line cap=round]
    (7.08,1.84) -- (7.08,1.61)
    (6.95,1.71) -- (7.08,1.80) -- (7.21,1.71)
    (6.96,1.39) -- (7.08,1.61) -- (7.20,1.39);
  \draw[OursColor,line width=.85pt] (5.44,1.28) rectangle (6.73,2.04);
  \draw[OursColor,line width=.85pt] (6.84,1.28) rectangle (7.35,2.10);
  \node[lbl] at (.87,.78)  {event packet\\(\(N\) tokens)};
  \node[lbl] at (3.27,.78) {mix + gate\\(on \(S<N\) states)};
  \node[lbl] at (6.40,.78) {spatial predictions\\(\(N\) features)};
  \node[font=\sffamily\bfseries\fontsize{7.5}{8.5}\selectfont,text=OursColor] at (4.15,.22)
    {\(+\)\, rich spatial context: high accuracy \qquad \(+\)\, low energy cost};
\end{tikzpicture}%

%% file: sec/0_abstract.tex
\begin{abstract}
% 1) Motivation: what the sensor produces and why detection with it matters.
Event cameras record brightness changes asynchronously, as a sparse stream of
events. They offer high temporal resolution and low power consumption, which
makes them a natural sensor for detecting moving objects.
% 2) Challenge: why it is hard.
But accurately detecting objects in sparse events data is compute expensive.
% 3) Trends: how others approach it, and what is missing.
To address this problem, many methods first group the events from a short time interval into packets, i.e., image-like tensors that show where and when activity occurred.
Processing packets densely keeps rich spatial context and thus gives high accuracy, but exchanging information across all spatial locations at every layer makes the energy cost high. Event-by-event
methods instead update only a small part of the network per event. This keeps
the energy cost low, but sacrifices accuracy.
% 4) Method: our approach.
We combine the best of both worlds, achieving high accuracy and low energy cost. Our detector processes event packets, but
first compresses their many spatial tokens into a few compact state-space model
(SSM) hidden states. The main feature mixing runs only on these states, and the
result is expanded back to every location to find objects.
% 5) Results.
On three public benchmarks (Gen1, N-Caltech101 and DSEC), our detector improves accuracy over the strongest asynchronous baselines from 0.72 to 9.92 AP points, while using 20$\times$ less computation and estimated energy.
% 6) Outlook.
This shows that compact SSM hidden states make accurate event-based detection
energy-efficient.
\end{abstract}

%% file: sec/1_intro.tex
\section{Introduction}
\label{sec:intro}

% 1) What is the problem?
Standard cameras, like those used in automotive, record a new image only every 33\,ms. Between two frames, a car driving at 50\,km/h moves about 0.4\,m,
and a pedestrian can step onto the road, unseen until the next frame
arrives~\cite{gehrig2024dagr}. Event cameras close this gap. Instead of
recording frames, each pixel independently reports a change in brightness the
moment it occurs as an event, with a position, a timestamp, and a polarity (the
sign of the change). Detecting objects with an event camera therefore means
turning this asynchronous stream of events into bounding boxes that say where
each object is and what it is.

% 2) Why is it important?
This sensing principle gives event cameras submillisecond latency, microsecond temporal resolution,
high dynamic range, and low power consumption. They see fast motion without
blur and keep working when a scene is too dark or too bright for standard
cameras~\cite{gehrig2024dagr}. These strengths, however, only matter if the
detector can keep up. A detector that is accurate but expensive cannot run
often enough on a robot or vehicle with a limited energy budget. A detector
that is cheap but inaccurate misses the objects that matter. Event-based
detection is only useful when it is both accurate and efficient.

% 3) Why is the problem hard?
This problem is difficult because a single event carries little information. Objects
only become visible when many events are combined across space. Gathering this
spatial context makes a detector accurate, but also expensive.

% 4) How far has existing work come? What is the next frontier?
Existing work follows two directions, shown in Fig.~\ref{fig:teaser}.
Packet-based methods group the events from a short time window into a packet:
an image-like tensor that records where and when activity
occurred~\cite{perot2020red,gehrig2023rvt,fan2025eventpillars}. A standard
network then processes this tensor densely. This keeps rich spatial context
and gives high accuracy, but the network mixes features at every spatial
location, so the energy cost is high. Event-by-event methods instead update
only the part of the network affected by each new event, using
graphs~\cite{schaefer2022aegnn,gehrig2024dagr} or sparse linear
attention~\cite{hao2026ssla}. This keeps the energy cost per event low, but
their accuracy remains below that of leading packet-based models. Moreover, when an application needs a result only once per packet,
event-by-event processing cannot recover the latency already spent waiting
for the packet, so processing the packet as a whole is the natural choice. The next frontier is therefore to process packets not only accurately, but
also at the amortized energy cost of event-by-event updates.

% 5) Why hasn't it been solved? What is the stumbling block?
The stumbling block is channel mixing. In a packet-based network, much of the
work consists of projections and gates applied to the feature vector at every
spatial location, so this work grows with the size of the packet. Faster
spatial mixing, such as linear-time attention or SSMs, does not remove it,
because every location is still transformed. One could avoid it by pooling the
whole packet into a single vector, as image classifiers do with global average
pooling. But such a vector only tells us what is in the scene, not where it
is, so it cannot be used to place bounding boxes with high accuracy.

% 6) What does our paper contribute?
We present a family of packet-based event detectors built on compact
state-space model (SSM) hidden states. We start from the hidden-state mixer of EfficientViM~\cite{lee2025efficientvim},
which builds on Mamba-2's state-space duality~\cite{dao2024mamba2}, and adapt
it to event packets. We combine it with
event statistics, multiscale prediction and task-specific training. For
RGB--event detection on DSEC \cite{gehrig2021dsec}, we also introduce conditional inference. The full
detector runs when a new RGB frame arrives and saves its temporal state. A
cheaper branch answers queries between frames by reusing the RGB features and
the saved state, without changing it. Each query thus uses the newest events
without running the full detector.

% 7) What is the key idea?
The key idea is to first compress the information, then mix it. Each mixer gathers the $N$
spatial tokens of a packet into a few hidden states ($S<N$), runs its main
channel mixing and gating on these states, and then expands the result back to
all $N$ locations. These transformations therefore act on $S$ vectors instead
of $N$, while the detector still sees a full spatial map to locate objects.
The packet keeps its rich spatial context, but
the cost of mixing no longer grows with the number of locations.

% 8) What do the experiments say?
On Gen1, our best detector reaches 39.65 AP, which is 2.15 points above SSLA-L
and 7.55 points above DAGR-L. All four model sizes exceed SSLA-L. On
N-Caltech101 Detection, it reaches 61.42 AP, 9.92 points above SSLA-L, while
using 17.15$\times$ less computation and 18.52$\times$ less estimated
energy~\cite{hao2026ssla}. On DSEC, our RGB--event detector improves endpoint
and interframe AP over DAGR by 1.13 and 0.72 points. It needs 0.56
MFLOPs and 0.45~$\mu$J per event, which is 11.48$\times$ and
12.02$\times$ less than DAGR's published figures, and its computation is
also below the 0.57 MFLOPs/event of DAGR's cheaper directed
variant~\cite{gehrig2024dagr}.\footnote{Ratios compare our amortized costs
with published reference figures. DSEC accounting covers complete streams at
about 192 queries/s. N-Caltech101 uses one prediction per clip. Energy is
estimated at 1.69 pJ/MAC.}
The same backbone also classifies events, with 91.91\% and 69.68\% top-1
accuracy on N-Caltech101 and N-ImageNet. On Gen1, pretraining adds 7.02 AP
over training from scratch, and polarity adds 1.58 AP.

Our contributions are:
\begin{itemize}
  \item A family of packet-based event detectors that compress spatial tokens
  into compact SSM hidden states before channel mixing. This gives high
  accuracy at a low computation and energy cost.
  \item Conditional SSM inference for RGB--event detection, which keeps full
  detections when RGB frames arrive and answers frequent queries between them
  with a cheaper branch.
  \item Four model sizes evaluated on three public detection benchmarks,
  supported by classification results and pretraining and polarity ablations.
\end{itemize}

We report DSEC semantic segmentation experiments in the supplementary material.

%% file: sec/2_related.tex
\section{Related Work}
\label{sec:related}

\paragraph{Event representations and packet-based perception.}
Aggregating events into spatial tensors makes their irregular measurements
compatible with established visual recognition architectures. Event Spike
Tensors provide a differentiable framework for learning this aggregation
jointly with a downstream task~\cite{gehrig2019est}. ERGO-12 selects
informative representation channels using a Gromov--Wasserstein discrepancy
between raw events and their encoded features~\cite{zubic2023ergo}.
EventPillars
encodes temporal range, polarity and normalized event density in a compact
representation, reducing encoding costs while retaining useful scene
structure~\cite{fan2025eventpillars}. For detection, RED combines dense event
representations with convolutional features and recurrent temporal
memory~\cite{perot2020red}. RVT extends this approach with local and global
spatial attention and recurrent feature aggregation~\cite{gehrig2023rvt}.
We retain packet statistics and multiscale features, and reduce the cost
after encoding by moving the mixer's main projections and gating into
compact SSM hidden states.

\paragraph{Asynchronous and sparse inference.}
A complementary direction reduces the work triggered by each arriving event.
AEGNN maintains a spatiotemporal graph and updates activations only at nodes
affected by new measurements~\cite{schaefer2022aegnn}. FARSE-CNN combines
recurrent processing with sparse convolutions and hierarchical compression
to process asynchronous event streams~\cite{santambrogio2024farse}.
DAGR combines an image CNN with an asynchronous event GNN, injecting image
context into the event branch to support predictions between image
arrivals~\cite{gehrig2024dagr}. SSLA partitions linear-attention state into
spatially selective components and combines sparse recurrent updates with
parallel scatter--compute--gather training~\cite{hao2026ssla}. Spiking
detectors, including MSD, pursue efficiency through spiking feature extraction
and multiscale detection~\cite{li2025msd}. These approaches exploit different
forms of sparsity and state reuse. We study an additional route: lowering the
cost of processing event packets through compact-state computation. Our DSEC
detector also reuses RGB features and committed temporal state, while
retaining current-event evidence at each scheduled query.

\paragraph{State-space models for visual and event processing.}
Mamba introduces input-dependent selection in state-space sequence models,
with computation that scales linearly with sequence
length~\cite{gu2024mamba, soydan2024s7, zubic2026regularity, zubic2026expressive}. Mamba-2 develops state-space duality, connecting
structured SSMs and attention and enabling efficient mixing
algorithms~\cite{dao2024mamba2, zubic2025gg}. Vision Mamba adapts selective SSMs to images
through bidirectional sequence processing~\cite{zhu2024visionmamba, raju2025perturbed}.
Earlier event-camera models combine temporal SSMs with spatial attention and learnable timescales to improve robustness to
changes in inference frequency~\cite{zubic2024eventssm}. SMamba exploits continuity
for token sparsification, combines spatial and channel interactions, and uses
ConvLSTM modules for temporal memory~\cite{yang2025smamba}. FocusMamba uses
motion information for collaborative RGB--event token sparsification and
fusion~\cite{yang2025focusmamba}. Building on compact hidden-state
mixing~\cite{lee2025efficientvim}, we adapt it to packet-based event detection.
Spatial tokens are aggregated before the mixer's main channel projections and gating,
then reconstructed into features for localization. This retains spatial
evidence from event packets while enabling accurate detection at low
amortized computation and MAC-energy cost.

\paragraph{Multimodal alignment, routing, and supervision.}
RGB--event fusion must reconcile fresh event measurements with appearance
features from an earlier frame. FAOD learns event-guided alignment from
temporally shifted training pairs~\cite{zhang2024faod}. ACGR aligns frame and
event graphs into a shared representation~\cite{li2025acgr}. HDI-Former
couples an image ANN and an event SNN through dynamic cross-modal
interaction~\cite{li2026hdiformer}. Adaptive fusion and temporally refined
pseudo-labels further support detection across operational
frequencies~\cite{lu2025flexevent}. These approaches motivate treating
multimodal alignment, query routing, and supervision as distinct design
choices. Our conditional SSM detector routes acquisition-boundary queries
through the full prediction path and interior queries through a cheaper
branch. Each interior query uses causal event prefixes and previews committed
memory. State is committed once per acquisition. This separates query
frequency from memory updates and reduces repeated inference work.

\paragraph{Pretraining and transfer for event vision.}
\vspace{-10px}
Limited event supervision makes initialization and representation adaptation
important \cite{lu2026hybrid}. N-ImageNet demonstrates the benefits of large-scale event
pretraining for recognition and learning with few labeled
examples~\cite{kim2021nimagenet}. Spatiotemporal prompting transfers
image-pretrained knowledge by adapting sparse event representations while
preserving their spatial and temporal structure~\cite{liang2025stp}.
Our event-vision family combines pretrained SSM features with task-specific
event statistics and detection heads. The pretraining and polarity ablations
examine these contributions separately from the choice of mixer. Together
with the classification evaluations, they test how compact SSM features
transfer across event tasks, while detection establishes the central
accuracy--computation and MAC-energy comparison.

%% file: sec/3_method.tex
\section{Method}
\label{sec:method}

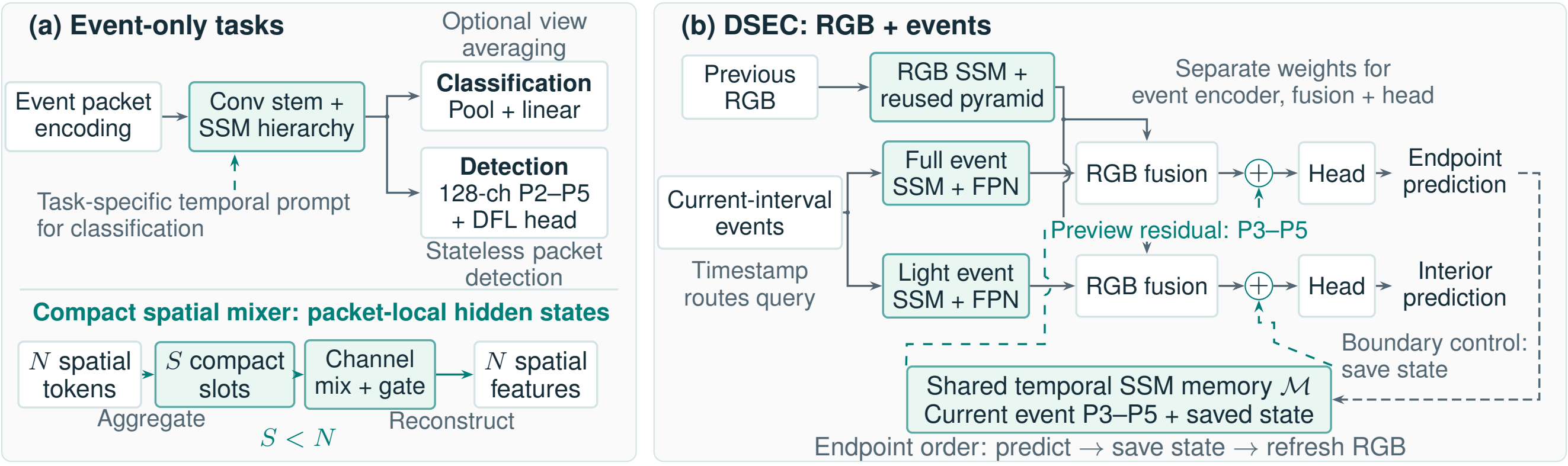
\begin{figure*}[t]
  \centering
  \resizebox{\textwidth}{!}{\input{figures/architecture}}
  \caption{\textbf{Compact SSM architectures for event vision.}
  \textbf{(a)} A convolutional stem and SSM stages turn encoded event packets
  into spatial features. Pooling supports classification, while a multiscale
  head predicts object boxes for Gen1 and N-Caltech101 Detection.
  The lower diagram shows tokens gathered into
  compact hidden states for channel mixing, then reconstructed as spatial features.
  \textbf{(b)} DSEC combines current events with reused features from the
  previous RGB frame. The full path predicts at frame boundaries, saves
  temporal state, then refreshes RGB features. The lighter path predicts
  between boundaries without changing the saved state.}
  \label{fig:architecture}
\end{figure*}

Our architecture retains the spatial context of event packets while reducing
the cost of global mixing. Figure~\ref{fig:architecture} shows its two uses:
event-only recognition and detection, and causal RGB--event detection with
frequent queries. Both use compact SSM hidden states. The latter additionally
reuses image features and temporal memory across queries.

\subsection{From event packets to compact hidden states}
\label{sec:compact_states}

An event $e_i=(x_i,y_i,t_i,p_i)$ records a pixel's brightness change at
time $t_i$ with polarity $p_i$. We group events into five temporal bins.
Each stores the first and last event times, polarity balance, and normalized
event counts~\cite{fan2025eventpillars}. These 20 channels describe when, where,
and how much activity occurred. Empty cells are zero. Gen1 and N-Caltech101
Detection bin at most 50,000 recent events over their time span.
DSEC uses fixed 10-ms bins starting at the previous RGB timestamp,
allowing successive queries to reuse accumulated statistics.

A convolutional stem feeds three spatial SSM stages. For event-only detection,
S, B, and M use two blocks per stage (six total), while L uses $(3,4,2)$
blocks (nine total). Each block contains one mixer with its own parameters.
Every mixer gathers spatial tokens into compact states, transforms them,
and reconstructs spatial features.

Within each mixer, $X\in\mathbb{R}^{C\times N}$ contains $N$ spatial
tokens with $C$ channels. A pointwise projection and depthwise convolution
predict write, read, and normalization maps
$B,R,\Delta\in\mathbb{R}^{S\times N}$. Following compact state-space
mixing~\cite{dao2024mamba2,lee2025efficientvim}, we compute
\begin{equation}
\begin{aligned}
 A &= \operatorname{softmax}_{N}(\Delta),\quad H=X(A\odot B)^\top, \\
 (U,Z)&=P_h(H), \\
 Y&=P_o\!\left(U\odot\operatorname{SiLU}(Z)+dU\right)R .
\end{aligned}
\label{eq:hidden_mixer}
\end{equation}
Here $H$ contains $S$ hidden states. The projections $P_h,P_o$ and gate
transform their channels, the learned scalar $d$ scales a residual contribution, and $R$
reconstructs spatial features for localization.

The saving comes from transforming fewer vectors. At fixed expansion
ratio, a channel projection costs $O(C^2)$ per vector. Applying it to
$S$ hidden states instead of $N$ spatial tokens changes its cost from
$O(NC^2)$ to $O(SC^2)$, reducing computation when $S<N$. Gathering and reconstructing spatial features
costs $O(NCS)$. Depthwise convolutions and feed-forward layers still
operate spatially. These hidden states are recomputed for each packet,
distinct from the persistent temporal memory below.

\subsection{Event-only prediction}
\label{sec:event_only}

For N-Caltech101 classification and event-only detection, EventPillars
packet encoding in Figure~\ref{fig:architecture}(a) converts temporal,
polarity, and density statistics into a spatial tensor. The convolutional
stem extracts local patterns, and the SSM hierarchy mixes information
across locations. The lower diagram illustrates its compact-state mixer
(Sec.~\ref{sec:compact_states}). Task-specific heads produce predictions.

For classification, pooling averages each stage's hidden states and the
final spatial features into summary vectors. Linear classifiers turn these
vectors into class scores, which we combine. N-Caltech101 uses one encoded recording.
N-ImageNet instead uses three overlapping 30-ms views, each encoded as five
bins of positive and negative event counts. Its temporal SSM prompt is an
input adapter, a trainable module that converts event bins into features
the pretrained backbone can use. It processes the bins in time order,
using a hidden state to retain earlier information. Residual SSM adapters
then add learned corrections to the backbone features. The same network
processes all views, and averaging their class scores produces one prediction.

For detection, spatial locations are retained to predict object positions.
A feature pyramid combines fine and coarse features into four maps,
called P2, P3, P4, and P5. Their height and width are reduced from the
padded input by factors of 4, 8, 16, and 32, respectively. Each has 128 channels.
Fine maps preserve small-object detail, while coarse maps provide broader
context. A shared head predicts class scores. Its box branch learns four
edge-distance distributions using distribution focal loss (DFL).
Their probability-weighted means define box edges relative to grid-cell
centers. Nonmaximum suppression removes overlapping duplicates.
As noted above, Gen1 and N-Caltech101 Detection process packets independently,
carrying no detector memory between them. This is the stateless path in the figure.
The separate DSEC event-only experiment adds temporal SSM processing of the
current packet and two preceding packets, as described in Sec.~\ref{sec:detection_results}.

\subsection{Causal RGB--event detection}
\label{sec:conditional_detection}

Figure~\ref{fig:architecture}(b) adds RGB appearance and temporal memory.
Let RGB frames arrive at $t_k$. A query $q\in(t_k,t_{k+1}]$ uses the
previous image $I_k$ and events with $t_k\le t_i<q$.
The RGB SSM computes its pyramid once per frame, then reuses these features
between arrivals. At a frame-arrival boundary, the full event path predicts,
saves memory, and refreshes RGB features for the next interval.
Queries between boundaries use a lighter private path with its own event
encoder, fusion, and head. They form temporary memory updates without
saving them, so additional queries cannot change subsequent predictions.
Empty and unlabelled intervals also execute boundary updates.

At each pyramid level, a learned gate combines shared RGB--event features
with features specific to each modality. RGB age and event activity guide
their relative weights. The RGB-only terms are reused, while event-dependent
terms are recomputed for each query.

Temporal memory consists of two SSM banks at P3--P5. Each bank uses
32, 48, and 64 SSM state channels per spatial location at these respective
levels. Every channel has four memory components with learned decay times,
initialized to 10, 50, 250, and 1,000 ms. Short decay times allow quick
responses to new events, while long ones retain older context.
From the last saved state $h_k$, each component computes
\begin{equation}
 \widetilde h_q=e^{-(q-t_k)/\tau}\odot h_k
       +\left(1-e^{-(q-t_k)/\tau}\right)\odot u(E_q),
 \label{eq:temporal_preview}
\end{equation}
where $\tau$ is its decay time and $u(E_q)$ transforms the current event
features into new evidence. Learned readouts turn both banks' states into
corrections added to the fused features. Only the full boundary path saves
$h_{k+1}=\widetilde h_{t_{k+1}}$.

Both heads predict distances from each P2--P5 grid point to the four box
edges. Each distance uses 17 bins numbered $0$--$16$, hence
$\texttt{reg\_max}=16$. Adjacent bins are one pyramid stride apart,
for example eight pixels at P3. The head averages these candidate distances
using their predicted probabilities as weights.

The lighter path keeps 128-channel outputs but uses fewer blocks and narrower
intermediate layers. Compressed projections mix features through fewer
channels. When input and output widths match, a scaled copy of each input
channel is also added. Spatial compression uses three linear convolutions:
$1\times1$ reduces channels, $3\times3$ filters neighbouring locations at
this smaller width, and $1\times1$ restores 128 channels. The expensive
spatial filtering therefore operates on fewer channels. A parallel
$3\times3$ filter processes each channel separately, and the two outputs
are added. Only this lighter path varies across DSEC's four sizes.
The full endpoint detector and both memory banks stay fixed.

\Needspace{6\baselineskip}
\subsection{Learning and amortized execution}
\label{sec:learning_execution}

Varifocal loss \cite{zhang2021varifocalnet} learns class scores reflecting box quality, GIoU \cite{rezatofighi2019giou} improves
box overlap, and DFL \cite{li2020gfl} learns edge-distance distributions. Their weights
are $1$, $2.5$, and $0.5$.

For DSEC, the frozen full detector teaches the lighter branch using the
same available events, RGB frame, and saved history. Cross-head
distillation~\cite{wang2024crosskd} feeds student features after the first
head block through the teacher's remaining layers. Matching their predictions
and internal features guides learning without adding inference work.
Boxes are imitated only when the teacher localizes better against ground truth.

Between annotated frames, we interpolate boxes along matching object tracks
and supervise only their known classes. Unknown locations and classes receive
no negative labels. Fully labelled endpoints also teach the lighter branch
to reject background.

We count all DSEC inference work: RGB refreshes, encoding, every query,
memory updates, and postprocessing. With $M$ multiply--accumulates (MACs),
$A$ scalar operations, and $N_e$ unique filtered events, amortized cost is
$(2M+A)/(10^6N_e)$ MFLOPs/event. Each MAC counts as two FLOPs.
At 1.69 pJ/MAC, the energy proxy is
$1.69M/(10^6N_e)$~$\mu$J/event. These averages charge all repeated work
while counting each input event once.

%% file: figures/architecture.tex
% Vector architecture figure. Load figures/teaser_style.tex before this file.
% Spatial HSM slots are local to one packet; M is persistent DSEC memory.
\begin{tikzpicture}[
  x=1cm,y=1cm,
  font=\sffamily\fontsize{7.6}{8.5}\selectfont,
  text=TeaserInk,
  flow/.style={-{Stealth[length=1.35mm,width=1.0mm]},line width=.65pt,draw=TeaserMuted},
  preview/.style={flow,dashed,draw=TeaserTeal},
  control/.style={flow,dash pattern=on 3pt off 1.5pt,draw=TeaserMuted},
  box/.style={draw=TeaserRule,fill=white,rounded corners=1.6pt,
    line width=.65pt,align=center,inner sep=3pt},
  ssm/.style={box,draw=TeaserTeal!65,fill=TeaserLight},
  label/.style={font=\sffamily\fontsize{7.5}{8.5}\selectfont,text=TeaserMuted,align=center},
  title/.style={font=\sffamily\bfseries\fontsize{8.6}{9.5}\selectfont},
  smalltitle/.style={font=\sffamily\bfseries\fontsize{7.7}{8.5}\selectfont},
  sum/.style={circle,draw=TeaserTeal,fill=white,inner sep=0pt,
    minimum size=.27cm,font=\sffamily\fontsize{8}{8}\selectfont}
]
  \path[use as bounding box] (0,0) rectangle (17.46,5.16);
  \filldraw[fill=TeaserMuted!3,draw=TeaserRule,rounded corners=3pt]
    (0,.02) rectangle (7.08,5.14);
  \filldraw[fill=TeaserMuted!3,draw=TeaserRule,rounded corners=3pt]
    (7.28,.02) rectangle (17.46,5.14);
  \node[title,anchor=west] at (.18,4.87) {(a) Event-only tasks};
  \node[title,anchor=west] at (7.46,4.87) {(b) DSEC: RGB + events};

  % Task-specific input encoders feed the compact spatial SSM hierarchy.
  \node[box,minimum width=1.30cm,minimum height=.77cm] (encoding)
    at (.91,3.87) {Event packet\\encoding};
  \node[ssm,minimum width=1.83cm,minimum height=.77cm] (hierarchy)
    at (3.07,3.87) {Conv stem +\\SSM hierarchy};
  \draw[flow] (encoding) -- (hierarchy);
  \node[box,minimum width=2.08cm,minimum height=.77cm] (class)
    at (5.72,4.10) {\textbf{Classification}\\Pool + linear};
  \node[box,minimum width=2.08cm,minimum height=.84cm] (detect)
    at (5.72,3.02) {\textbf{Detection}\\128-ch P2--P5\\+ DFL head};
  \draw[flow] (hierarchy.east) -- (4.30,3.87) |- (class.west);
  \draw[flow] (4.30,3.87) |- (detect.west);
  \fill[TeaserMuted] (4.30,3.87) circle (.025);
  \node[label,text width=2.08cm,inner sep=0pt] at (5.72,4.77)
    {Optional view\\averaging};
  \node[label,text width=2.08cm,inner sep=0pt] at (5.72,2.23)
    {Stateless packet\\detection};
  \node[label,align=left,anchor=west] at (.27,2.78)
    {Task-specific temporal prompt\\for classification};
  \draw[preview] (2.60,3.05) -- (2.60,3.45);

  % Shared compact hidden-state mixer, explicitly separated from temporal M.
  \draw[TeaserRule,line width=.6pt] (.20,1.94) -- (6.88,1.94);
  \node[smalltitle,anchor=west,text=TeaserTeal] at (.22,1.67)
    {Compact spatial mixer: packet-local hidden states};
  \node[box,font=\sffamily\fontsize{8}{9}\selectfont,minimum width=1.05cm,minimum height=.60cm] (tokens)
    at (.87,.99) {$N$ spatial\\tokens};
  \node[ssm,font=\sffamily\fontsize{8}{9}\selectfont,minimum width=1.13cm,minimum height=.60cm] (slots)
    at (2.49,.99) {$S$ compact\\slots};
  \node[ssm,minimum width=1.22cm,minimum height=.60cm] (mix)
    at (4.11,.99) {Channel\\mix + gate};
  \node[box,font=\sffamily\fontsize{8}{9}\selectfont,minimum width=1.33cm,minimum height=.60cm] (spatial)
    at (5.96,.99) {$N$ spatial\\features};
  \draw[flow,draw=TeaserTeal] (tokens) -- (slots);
  \draw[flow,draw=TeaserTeal] (slots) -- (mix);
  \draw[flow,draw=TeaserTeal] (mix) -- (spatial);
  \node[label] at (1.67,.48) {Aggregate};
  \node[label] at (5.02,.48) {Reconstruct};
  \node[label,font=\sffamily\fontsize{8}{9}\selectfont,text=TeaserTeal] at (3.31,.28) {$S<N$};

  % DSEC has separate full and interior branches. Both reuse the RGB pyramid.
  \node[box,minimum width=1.51cm,minimum height=.52cm] (rgb)
    at (8.35,4.20) {Previous\\RGB};
  \node[ssm,minimum width=2.05cm,minimum height=.52cm] (cache)
    at (10.72,4.20) {RGB SSM +\\reused pyramid};
  \draw[flow] (rgb) -- (cache);
  \node[box,minimum width=1.51cm,minimum height=.81cm] (events)
    at (8.35,2.80) {Current-interval\\events};
  \node[label] at (8.35,1.98) {Timestamp\\routes query};
  \node[ssm,minimum width=1.61cm,minimum height=.69cm] (full)
    at (10.65,3.24) {Full event\\SSM + FPN};
  \node[ssm,minimum width=1.61cm,minimum height=.69cm] (cheap)
    at (10.65,1.98) {Light event\\SSM + FPN};
  \draw[flow] (events.east) -- (9.44,2.80) |- (full.west);
  \draw[flow] (9.44,2.80) |- (cheap.west);
  \fill[TeaserMuted] (9.44,2.80) circle (.025);
  \node[box,minimum width=1.49cm,minimum height=.69cm] (fusionfull)
    at (12.78,3.24) {RGB fusion};
  \node[box,minimum width=1.49cm,minimum height=.69cm] (fusioncheap)
    at (12.78,1.98) {RGB fusion};
  \draw[flow] (full) -- (fusionfull);
  \draw[flow] (cheap) -- (fusioncheap);
  % A vertical RGB bus crosses the independent full event-feature edge.
  \draw[TeaserMuted,line width=.65pt] (cache.east) -- (11.85,4.20)
    -- (11.85,3.34);
  \draw[TeaserMuted,line width=.65pt] (11.85,3.34)
    arc[start angle=90,end angle=-90,x radius=.07cm,y radius=.10cm];
  \draw[flow] (11.85,3.14) -- (11.85,2.53) -- (12.78,2.53)
    -- (fusioncheap.north);
  \draw[flow] (11.85,3.84) -- (12.78,3.84) -- (fusionfull.north);
  \fill[TeaserMuted] (11.85,3.84) circle (.025);
  \node[sum] (sumfull) at (14.03,3.24) {$+$};
  \node[sum] (sumcheap) at (14.03,1.98) {$+$};
  \draw[flow] (fusionfull) -- (sumfull);
  \draw[flow] (fusioncheap) -- (sumcheap);
  \node[box,minimum width=.74cm,minimum height=.69cm] (headfull)
    at (14.90,3.24) {Head};
  \node[box,minimum width=.74cm,minimum height=.69cm] (headcheap)
    at (14.90,1.98) {Head};
  \draw[flow] (sumfull) -- (headfull);
  \draw[flow] (sumcheap) -- (headcheap);
  \node[align=center] (endpoint) at (16.22,3.24) {Endpoint\\prediction};
  \node[align=center] (interior) at (16.22,1.98) {Interior\\prediction};
  \draw[flow] (headfull) -- (endpoint);
  \draw[flow] (headcheap) -- (interior);
  \node[label] at (14.30,4.25) {Separate weights for\\event encoder, fusion + head};

  % Both branches preview shared saved memory. Only the full path saves state.
  \node[ssm,font=\sffamily\fontsize{8}{9}\selectfont,minimum width=4.73cm,minimum height=.64cm] (memory)
    at (12.47,.71) {Shared temporal SSM memory $\mathcal{M}$\\Current event P3--P5 + saved state};
  \draw[preview] (memory.north east) -- (14.20,1.35) -- (14.20,1.57)
    -- (14.03,1.57) -- (sumcheap.south);
  \draw[TeaserTeal,dashed,line width=.65pt] (memory.north west)
    -- (10.10,1.33) -- (11.64,1.33) -- (11.64,1.88);
  \draw[TeaserTeal,dashed,line width=.65pt] (11.64,1.88)
    arc[start angle=-90,end angle=90,x radius=.065cm,y radius=.10cm];
  \draw[preview] (11.64,2.08) -- (11.64,2.61)
    -- (14.03,2.61) -- (sumfull.south);
  \node[label,fill=TeaserMuted!3,inner sep=1pt,text=TeaserTeal]
    at (13.14,2.61) {Preview residual: P3--P5};
  % Dashed control edge sequences the save. Predictions are not state inputs.
  \draw[control] (endpoint.east) -- (17.16,3.24)
    -- (17.16,.71) -- (memory.east);
  \node[label,fill=TeaserMuted!3,inner sep=1pt,align=left]
    at (15.99,1.20) {Boundary control:\\save state};
  \node[label,font=\sffamily\fontsize{8}{9}\selectfont] at (12.37,.16)
    {Endpoint order: predict $\rightarrow$ save state $\rightarrow$ refresh RGB};
\end{tikzpicture}%

%% file: sec/2_experiments.tex
\section{Experiments}
\label{sec:experiments}

We test whether packet-based SSMs can improve detection accuracy while
keeping inference inexpensive. We first compare four model sizes with
asynchronous and dense detectors, then examine their computational costs
and the training choices that contribute to accuracy. We also evaluate
event classification to test whether the same compact SSM design can
recognize object categories from event packets.

\subsection{Experimental setup}
\label{sec:experimental_setup}

\textbf{Datasets and metrics.}
Gen1~\cite{deTournemire2020gen1} uses events alone to detect cars and pedestrians on the official test
split, with a warm-up of 0.5\,s and the standard small-box filtering.
N-Caltech101 Detection~\cite{orchard2015ncaltech} uses 906 held-out clips and 100 foreground classes,
with one prediction at the end of each clip. DSEC uses causal events,
either alone or with the previous RGB frame. RGB+event evaluation uses six
validation recordings and two classes, car and pedestrian. Endpoint AP covers 5,897 labelled frame
boundaries, while interframe AP covers 5,094 released intermediate targets.
The two scores use different annotated samples. Their difference therefore
does not directly measure whether detection becomes more or less accurate
between RGB frames. Detection AP averages IoU thresholds from
0.50 to 0.95. We report AP50 separately and express all accuracies in percent.

\textbf{Models and selection.}
For event-only detection, including DSEC, S/B/M/L denote four spatial SSM sizes.
Only RGB+event DSEC uses four lighter branches sharing one frozen
endpoint detector. Training follows Sec.~\ref{sec:learning_execution},
with pretrained SSM backbones, AdamW \cite{loshchilov2019adamw}, and validation-based selection.
Published baselines retain their reported evaluation protocols, identified in the tables.

\input{tables/experiments_event_only}

\subsection{Object detection}
\label{sec:detection_results}

\textbf{Gen1 automotive detection.}
Table~\ref{tab:event_detection} shows that every model size exceeds the
published 37.50 AP of SSLA-L. Our medium model reaches 39.65 AP and
69.39 AP50, improving AP by 2.15 points over SSLA-L and 7.55 points over
DAGR-L. Even S reaches 38.24 AP. These gains show that compact-state
processing preserves useful localization information at modest model
sizes. Increasing capacity is not uniformly beneficial: M outperforms L,
so we report the complete family rather than only its largest member.

The same table includes dense approaches from the published comparisons.
Our models exceed Inception+SSD, Events+RRC, MatrixLSTM+YOLOv3, and
Events+YOLOv3 in AP. Some dense systems achieve higher accuracy, including
EventPillars at 53.10 AP, but their reported computational costs are
substantially higher.

\textbf{N-Caltech101 Detection.}
The improvement also transfers beyond automotive scenes. Our M model
reaches 61.42 AP and 83.79 AP50, exceeding SSLA-L by 9.92 AP and 9.49 AP50
points. Its AP50 also exceeds DAGR-L by 10.59 points. Across corresponding
S/B/M/L sizes, AP gains over SSLA are 11.58, 10.15, 11.92, and 9.82 points.
The two detector seeds give mean AP and standard deviation of
$55.93\pm0.08$, $57.67\pm1.10$, $61.38\pm0.05$, and $61.25\pm0.10$,
respectively.

\input{tables/experiments_dsec}

\textbf{DSEC detection between RGB frames.}
We compare packet-based SSMs with asynchronous DAGR on the same validation targets.
L reaches 43.41 endpoint AP and 47.01 interframe AP, exceeding DAGR-R50
by 1.13 and 0.72 points (Table~\ref{tab:dsec_detection}(a)). All four
RGB+event variants use the same frozen endpoint detector and weights,
so their endpoint predictions and AP are identical. S/B/M/L vary only
the cheaper branch used between RGB frames. Interframe AP rises from 35.85 (S) to
47.01 (L), showing the benefit of retaining capacity for current-event
processing between RGB arrivals.

The added references retain their two-class test protocols. EFH-18
($\dagger$) reports AP at 25\,ms, while EFH-10, Events+YOLOX and
DAGR-R18/R34 ($\ddagger$) report peak AP over 0--50\,ms~\cite{zhang2026efh,gehrig2024dagr}.
SpikeYOLO8l uses CNN fusion and its native late-interframe evaluation
($\S$)~\cite{khitushkin2026spikeyolo}.
Our models and DAGR-R50 use common validation at approximately 180\,Hz.
$P$ marks published costs, and $*$ marks costs of EFH's SNN event branch only.
SpikeYOLO's per-event costs are unreported and remain blank.

\textbf{DSEC event-only detection.}
We train four event-only models at $430\times640$ resolution, with temporal
SSMs processing the current packet and two preceding 50\,ms packets.
We choose checkpoints and test-time augmentation (TTA) on validation, then
evaluate 12,131 held-out test frames. All four sizes exceed reported
SSLA-L in both metrics (Table~\ref{tab:dsec_detection}(b)). L reaches
36.08 AP and 63.21 AP50, gains of 9.28 and 8.01 points over SSLA-L,
and 17.18 and 25.11 points over event-only DAGR-L~\cite{hao2026ssla}.

\subsection{Efficiency and model scaling}
\label{sec:efficiency_results}

\input{figures/experiments_tradeoff}
\input{figures/experiments_memory}

For DSEC and N-Caltech101, complete-input cost includes event encoding,
detection, memory updates and final box processing. We divide total work
by unique filtered events. Repeated queries add computation without
counting their events twice. DSEC covers 526.11 seconds and 833.33 million
events at approximately 192 queries per second. N-Caltech101 covers
906 clips and 44.46 million events.

On DSEC, L requires 0.56 MFLOPs/event and 0.45~$\mu$J/event.
These are 11.48 and 12.02 times lower than regular DAGR's published costs,
and below directed DAGR's 0.57 MFLOPs/event and 0.96~$\mu$J/event.
Choosing S saves another 29.12\% of computation but loses 11.16 interframe
AP points. The shared endpoint detector retains its accuracy and cost.

On N-Caltech101, M reaches 83.79 AP50 at 0.05 MFLOPs/event and
0.04~$\mu$J/event (Fig.~\ref{fig:efficiency}). It exceeds SSLA-L by
9.49 AP50 points with 17.15 times less computation and 18.52 times less
MAC-based energy.

Figure~\ref{fig:memory} compares peak GPU memory on Gen1 and N-Caltech101.
Across corresponding sizes, SSLA uses 2.06--4.09 times our inference
memory and 2.92--4.69 times our training memory. Our inference memory
is also lower than all four DAGR configurations. Training memory is
closer, with our L slightly higher than DAGR-L on both datasets.

These comparisons use the accounting scopes stated in the tables.
Gen1 reports the neural cost of one packet divided by its historical
mean of 49,291 events, excluding representation work and unsupported
operations. Energy is computed from MACs as defined in
Sec.~\ref{sec:learning_execution}, rather than measured at the GPU.

\subsection{Ablation studies}
\label{sec:ablation_results}

\input{tables/experiments_ablation}

Table~\ref{tab:ablations} tests which training and detection choices help
our M-sized SSM on Gen1 validation. The full recipe reaches 40.17 AP.
Each variant uses the same training limit and stopping rule.

The control starts from weights pretrained on N-ImageNet classification.
Starting from random weights instead loses 7.02 AP, showing the value of
pretraining at this budget. Starting directly from RGB-pretrained weights
changes AP by only $+0.10$. The uncertainty interval includes zero, so
these runs do not establish an advantage for event pretraining over RGB.

Removing random crops and translations loses 2.47 AP. To test polarity,
we zero its channels during training and evaluation while preserving event
times, counts and input dimensions. The 1.58 AP loss shows that the
direction of brightness changes supplies useful information. Replacing
the detection path with RetinaNet loses 1.16 AP. Its pyramid, head,
box matching and losses change together, while retaining the same SSM
backbone architecture.

Finally, we append four statistics from up to 10,000 events in the last
100\,ms. A $1\times1$ projection maps the resulting 24 channels back to
20 before the same backbone. This emphasizes recent activity already
in the packet and adds 0.54 AP at 2.08\% more counted network computation,
excluding event encoding. Resampling validation recordings gives a 95\%
interval of $[0.04,0.95]$ AP for the gain in these trained models.
This validation variant does not replace our reported test model.

\subsection{Event classification}
\label{sec:classification_results}

\input{tables/experiments_classification}

Table~\ref{tab:classification} tests whether the architecture also supports
recognition without box prediction. N-Caltech101 reaches 91.91\% top-1
on the complete 1,742-example holdout, exceeding the reported
MEM-NImNet result by 1.81 points. The five-bin model outperforms
its ten-bin counterpart by 0.52 points, so finer temporal partitioning is
not uniformly helpful. This classification split contains 101 classes and
is separate from the detection split.

On the official 50,000-example N-ImageNet validation set, three temporal
views through one shared model yield 69.68\% top-1 and 88.98\% top-5.
Top-1 exceeds PASS-M-K(2), EventBind-L and STP by 8.36, 6.14 and
0.81 points, respectively. STP retains higher top-5.

%% file: tables/experiments_event_only.tex
% Generated by scripts/build_event_only_table.py. Do not edit numeric cells manually.
\begin{table*}[t]
\centering
\caption{\textbf{Event-only detection on Gen1 and N-Caltech101.}
All four compact SSMs exceed SSLA-L in AP on both datasets. AP and AP$_{50}$ are percentages.
$F$ denotes computation in MFLOPs/event and $E$ denotes MAC energy in $\mu$J/event.
Bold and underlining mark the best and second-best listed values in each column, respectively, including ties.}
\label{tab:event_detection}
\begingroup\scriptsize
\setlength{\tabcolsep}{3pt}
\renewcommand{\arraystretch}{1.0}
\begin{tabular*}{\textwidth}{@{\extracolsep{\fill}}lrrrrrrrr@{}}
\toprule
& \multicolumn{4}{c}{\textbf{Gen1 Detection}} & \multicolumn{4}{c}{\textbf{N-Caltech101 Detection}}\\
\cmidrule(lr){2-5}\cmidrule(lr){6-9}
Method & AP $\uparrow$ & AP$_{50}$ $\uparrow$ & $F\downarrow$ & $E\downarrow$
& AP $\uparrow$ & AP$_{50}$ $\uparrow$ & $F\downarrow$ & $E\downarrow$\\
\midrule
\multicolumn{9}{l}{\textbf{Dense detectors ($P$)}}\\
Inception+SSD~\cite{iacono2018inceptionssd} & 30.10 & -- & 27,183 & 23,502 & -- & -- & -- & --\\
Events+RRC~\cite{chen2018eventsrrc} & 30.70 & -- & 21,758$^{*}$ & 18,386 & -- & -- & -- & --\\
MatrixLSTM+YOLOv3~\cite{cannici2020matrixlstm} & 31.00 & -- & 34,519$^{*}$ & 29,168 & -- & -- & -- & --\\
Events+YOLOv3~\cite{jiang2019eventsyolo} & 31.20 & -- & 65,558 & 55,397 & -- & -- & -- & --\\
RED~\cite{perot2020red} & 40.00 & -- & 4,712 & 3,982 & -- & -- & -- & --\\
S5-ViT-B~\cite{zubic2024eventssm} & 47.70 & -- & -- & -- & -- & -- & -- & --\\
EVA+RVT-B~\cite{hao2026eva} & 47.70 & -- & 3,500 & -- & -- & -- & -- & --\\
GET~\cite{peng2023get} & 47.90 & -- & 3,600 & -- & -- & -- & -- & --\\
SAST-CB~\cite{peng2024sast} & 48.20 & -- & 2,400 & -- & -- & -- & -- & --\\
SMamba~\cite{yang2025smamba} & \underline{50.40} & -- & 2,400 & -- & -- & -- & -- & --\\
ERGO-12~\cite{zubic2023ergo} & \underline{50.40} & -- & 50,800 & -- & -- & -- & -- & --\\
EventPillars~\cite{fan2025eventpillars} & \textbf{53.10} & -- & 50,800 & -- & -- & -- & -- & --\\
\midrule
\multicolumn{9}{l}{\textbf{Asynchronous detectors ($P$)}}\\
NVS-S~\cite{li2021nvss} & -- & 8.60 & 7.80 & 6.59 & -- & 34.60 & 7.80 & 6.59\\
AsyNet~\cite{messikommer2020asynet} & -- & 14.50 & 205.00 & 173.00 & -- & 64.30 & 200.00 & 169.00\\
AEGNN~\cite{schaefer2022aegnn} & -- & 16.30 & 5.26 & 4.44 & -- & 59.50 & 7.41 & 6.26\\
FARSE-CNN~\cite{santambrogio2024farse} & -- & 30.00 & 0.14 & -- & -- & -- & -- & --\\
EHGCN~\cite{chen2025ehgcn} & -- & -- & -- & -- & -- & 69.40 & 1.06 & --\\
\addlinespace[2pt]
DAGR-N~\cite{gehrig2024dagr} & 26.30 & -- & 1.36 & 1.15 & -- & 62.90 & 2.28 & 1.93\\
DAGR-S~\cite{gehrig2024dagr} & 30.40 & -- & 4.58 & 3.87 & -- & 70.20 & 6.85 & 5.76\\
DAGR-M~\cite{gehrig2024dagr} & 31.80 & -- & 9.94 & 8.40 & -- & 72.70 & 12.20 & 10.60\\
DAGR-L~\cite{gehrig2024dagr} & 32.10 & -- & 17.40 & 14.70 & -- & 73.20 & 18.90 & 16.00\\
\addlinespace[2pt]
SSLA-S~\cite{hao2026ssla} & 33.40 & 62.90 & 0.10 & 0.09$^{\dagger}$ & 44.40 & 68.10 & 0.13 & 0.11$^{\dagger}$\\
SSLA-B~\cite{hao2026ssla} & 35.10 & 65.50 & 0.18 & 0.15$^{\dagger}$ & 48.30 & 72.00 & 0.23 & 0.20$^{\dagger}$\\
SSLA-M~\cite{hao2026ssla} & 37.00 & 67.00 & 0.41 & 0.34$^{\dagger}$ & 49.50 & 72.40 & 0.52 & 0.44$^{\dagger}$\\
SSLA-L~\cite{hao2026ssla} & 37.50 & 67.50 & 0.72 & 0.61$^{\dagger}$ & 51.50 & 74.30 & 0.93 & 0.78$^{\dagger}$\\
\midrule
\multicolumn{9}{l}{\textbf{Our compact SSMs (Gen1 $W$ / N-Caltech101 $T$)}}\\
Ours S & 38.24 & 67.24 & \textbf{0.05} & \textbf{0.04} & 55.98 & 77.97 & \textbf{0.04} & \textbf{0.03}\\
Ours B & 37.73 & 66.84 & \underline{0.06} & \underline{0.05} & 58.45 & 80.61 & \underline{0.04} & \underline{0.03}\\
Ours M & 39.65 & \textbf{69.39} & 0.08 & 0.07 & \textbf{61.42} & \textbf{83.79} & 0.05 & 0.04\\
Ours L & 39.04 & \underline{67.66} & 0.09 & 0.08 & \underline{61.32} & \underline{82.93} & 0.06 & 0.05\\
\bottomrule
\end{tabular*}
\par\smallskip
\begin{minipage}{\textwidth}\scriptsize
\textit{Cost definitions.} $P$ retains published native costs from SSLA~\cite{hao2026ssla}
and DAGR~\cite{gehrig2024dagr}. $W$ counts supported neural-forward operations per historical
packet event, excluding representation construction and unsupported operations.
$T$ counts the complete terminal prediction per unique filtered event.
$^{*}$ marks a backbone-only lower bound. $^{\dagger}$ marks converted energy,
$E=0.845F$, using two FLOPs/MAC and 1.69 pJ/MAC. Dashes indicate unreported results.
\end{minipage}
\endgroup
\end{table*}

%% file: tables/experiments_dsec.tex
% Generated by scripts/build_dsec_assets.py from experiments/dsec.json.
\begin{table*}[t]
\centering
\footnotesize
\caption{\textbf{DSEC Detection.} (a) Two-class RGB+event detection.
(b) Our held-out test TTA results and reported event-only baselines~\cite{hao2026ssla}.
TTA combines flips and, for S, multiple scales, selected on validation.
AP/AP$_{50}$ are percentages. Costs apply to (a): $F$ is MFLOPs/event and $E$ is arithmetic $\mu$J/event. Markers are defined below.
Bold/underlining mark best/second-best; AP rankings in (a) cover common validation only.}
\label{tab:dsec_detection}
\begin{minipage}[t]{.62\textwidth}
\centering
\textbf{(a) RGB+event detection}\par\smallskip
\setlength{\tabcolsep}{3pt}
\begin{tabular}{lrrrr}
\toprule
Method & \shortstack{Endpoint\\AP $\uparrow$} & \shortstack{Interframe\\AP $\uparrow$} & $F\downarrow$ & $E\downarrow$\\
\midrule
EFH-10~\cite{zhang2026efh} & -- & 41.90$^{\ddagger}$ & 1.88$^{*}$ & 1.79$^{*}$\\
EFH-18~\cite{zhang2026efh} & -- & 45.10$^{\dagger}$ & 3.32$^{*}$ & 3.09$^{*}$\\
SpikeYOLO8l~\cite{khitushkin2026spikeyolo} & -- & 30.70$^{\S}$ & -- & --\\
Events+YOLOX~\cite{gehrig2024dagr} & 40.20 & 42.50$^{\ddagger}$ & 22,049$^{P}$ & 18,631$^{P}$\\
DAGR-R18~\cite{gehrig2024dagr} & 37.60 & 39.70$^{\ddagger}$ & 6.74$^{P}$ & 5.70$^{P}$\\
DAGR-R34~\cite{gehrig2024dagr} & 39.00 & 41.40$^{\ddagger}$ & 6.57$^{P}$ & 5.55$^{P}$\\
DAGR-R50~\cite{gehrig2024dagr} & \underline{42.28} & \underline{46.29} & 6.41$^{P}$ & 5.42$^{P}$\\
\midrule
Ours S & \textbf{43.41} & 35.85 & \textbf{0.39} & \textbf{0.31}\\
Ours B & \textbf{43.41} & 42.93 & \underline{0.44} & \underline{0.35}\\
Ours M & \textbf{43.41} & 45.54 & 0.49 & 0.39\\
Ours L & \textbf{43.41} & \textbf{47.01} & 0.56 & 0.45\\
\bottomrule
\end{tabular}
\par\smallskip
\begin{minipage}{\linewidth}\scriptsize
DAGR-R50 and ours: common validation. Other rows: native test AP,
$\dagger$25\,ms, $\ddagger$peak over 0--50\,ms, $\S$late interframe.
$P$: published cost. $*$: SNN event branch only.
Ours share a frozen endpoint detector and report complete dense-stream amortized costs.
\end{minipage}
\end{minipage}\hfill
\begin{minipage}[t]{.36\textwidth}
\centering
\textbf{(b) Event-only detection}\par\smallskip
\setlength{\tabcolsep}{3pt}
\begin{tabular}{lrr}
\toprule
Method & AP $\uparrow$ & AP$_{50}$ $\uparrow$\\
\midrule
FARSE-CNN~\cite{santambrogio2024farse} & 0.4 & 1.3\\
DAGR-N~\cite{gehrig2024dagr} & 16.2 & 34.7\\
DAGR-S~\cite{gehrig2024dagr} & 17.8 & 36.2\\
DAGR-M~\cite{gehrig2024dagr} & 18.8 & 38.0\\
DAGR-L~\cite{gehrig2024dagr} & 18.9 & 38.1\\
\midrule
SSLA-S~\cite{hao2026ssla} & 22.9 & 49.6\\
SSLA-B~\cite{hao2026ssla} & 23.4 & 50.6\\
SSLA-M~\cite{hao2026ssla} & 26.2 & 54.6\\
SSLA-L~\cite{hao2026ssla} & 26.8 & 55.2\\
\midrule
Ours S & 33.01 & 58.38\\
Ours B & 33.06 & 58.41\\
Ours M & \underline{35.56} & \underline{62.62}\\
Ours L & \textbf{36.08} & \textbf{63.21}\\
\bottomrule
\end{tabular}
\end{minipage}
\end{table*}

%% file: figures/experiments_tradeoff.tex
% Generated by scripts/build_dsec_assets.py.
\begin{figure}[t]
\centering
\includegraphics[width=\columnwidth]{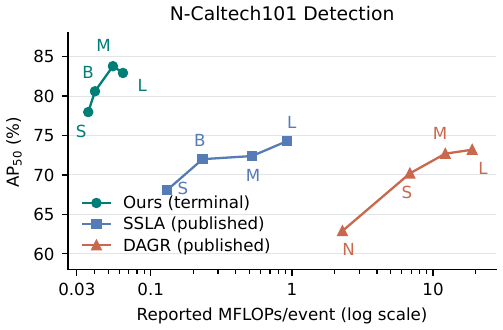}
\caption{\textbf{Accuracy and computation on N-Caltech101 Detection.}
Each point represents one model size. Our models combine higher AP50
with lower computation per event. Our costs cover one complete prediction
per clip, while SSLA and DAGR retain their published incremental costs.}
\label{fig:efficiency}
\end{figure}

%% file: figures/experiments_memory.tex
% Generated by scripts/build_memory_assets.py from completed GPU receipts.
\begin{figure}[t]
\centering
\includegraphics[width=\columnwidth]{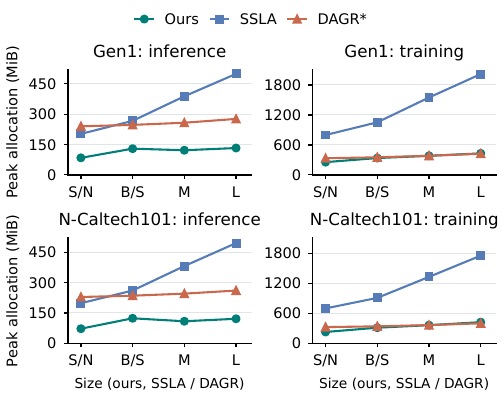}
\caption{\textbf{GPU memory on Gen1 and N-Caltech101 Detection.}
Peak FP32 tensor allocation at batch size one on eight shared clips per
dataset, capped at 50k events. Training includes gradients and AdamW,
inference includes decoding and NMS. Sizes are S/B/M/L for ours and SSLA,
N/S/M/L for DAGR. $*$DAGR uses dataset-adapted released architectures
with random weights. These Python packet profiles exclude CPU preprocessing,
teachers and EMA, and do not measure incremental deployment.}
\label{fig:memory}
\end{figure}

%% file: tables/experiments_ablation.tex
% Generated by scripts/build_support_tables.py from verified confirmation runs.
\begin{table}[t]
\centering
\footnotesize
\caption{\textbf{What matters on Gen1?} Validation AP for model M (mean $\pm$ SD over three seeds) and paired changes from the control. These seven configurations were confirmed after screening.}
\label{tab:ablations}
\setlength{\tabcolsep}{3pt}
\begin{tabular}{lrr}
\hline
Change from control & AP $\pm$ SD & $\Delta$AP\\
\hline
Full control & $40.17\pm0.14$ & $0.00$\\
Direct RGB initialization & $40.27\pm0.10$ & $+0.10$\\
Training from scratch & $33.15\pm0.19$ & $-7.02$\\
RetinaNet \cite{lin2017retinanet} package & $39.01\pm0.27$ & $-1.16$\\
No crop/translation & $37.70\pm0.44$ & $-2.47$\\
Mask average polarity & $38.59\pm0.36$ & $-1.58$\\
Add recent-event context & $\mathbf{40.71\pm0.39}$ & $+0.54$\\
\hline
\end{tabular}
\end{table}

%% file: tables/experiments_classification.tex
% Generated by scripts/build_support_tables.py. Published protocols retained.
\begin{table}[t]
\centering
\footnotesize
\caption{\textbf{Event classification.} Top-1/Top-5 accuracy (\%) against selected published references with their original protocols. Ours uses a 1,742-example N-Caltech101 holdout and three 256px temporal views on official N-ImageNet validation. Bold marks the best listed score per dataset. Dashes indicate unreported values.}
\label{tab:classification}
\setlength{\tabcolsep}{4pt}
\begin{tabular}{lrr}
\hline
Method & Top-1 & Top-5\\
\hline
\multicolumn{3}{l}{\textit{N-Caltech101}}\\
EST~\cite{gehrig2019est} & 81.70 & --\\
E2VID~\cite{rebecq2019e2vid} & 86.60 & --\\
ECDP~\cite{yang2023ecdp} & 87.66 & --\\
MEM-NImNet~\cite{klenk2024mem} & 90.10 & --\\
Video-to-Events~\cite{gehrig2020videoevents} & 90.60 & --\\
\textbf{Ours} & \textbf{91.91} & \textbf{97.70}\\
\hline
\multicolumn{3}{l}{\textit{N-ImageNet}}\\
MEM~\cite{klenk2024mem} & 57.89 & --\\
PASS-M-K(2)~\cite{zhou2025pass} & 61.32 & --\\
EventBind (ViT-L/14)~\cite{zhou2024eventbind} & 63.54 & --\\
ECDP~\cite{yang2023ecdp} & 64.83 & 86.30\\
STP~\cite{liang2025stp} & 68.87 & \textbf{89.65}\\
\textbf{Ours} & \textbf{69.68} & 88.98\\
\hline
\end{tabular}
\end{table}

%% file: sec/5_conclusion.tex
\section{Conclusion}
\label{sec:conclusion}

We show that event packets can support accurate object detection at low
computational cost through compact SSM hidden states. The mixer's main
channel transformations act on a small set of states, while reconstructed
spatial features provide the detail needed to locate objects. For RGB--event
detection, a cheaper branch reuses image features and saved temporal state
between full predictions at image arrivals. Across Gen1, N-Caltech101
Detection and DSEC, our detectors improve accuracy over strong asynchronous
baselines. On DSEC and N-Caltech101 Detection, these gains accompany lower
amortized computation and MAC energy. These results demonstrate that
packet-based spatial context can be retained within an efficient SSM detector.

%% file: sec/6_LLM_usage.tex
\section*{LLM Usage Statement}

% disclose any use of LLMs or other generative AI tools in the research or the writing, including for editing and grammar;
% name the tool(s) used, identify the parts of the paper or the research affected (for example text, figures, code, experiments, related work or ideation), and briefly describe the nature and extent of the use. If naming a non-public or organization-specific tool would compromise anonymity, authors should provide a generic technical description during review and provide the full tool identification in the camera-ready version; and
% state “No LLMs or generative AI tools were used in this work” if none were used.

We used GPT-6 Astra to assist with code implementation and manuscript revision, and to obtain simulated reviewer feedback. The authors checked and revised all code contributed by the model for correctness and consistency with the proposed method. The research ideas, experimental design, and analysis originated with the authors. References were added manually without LLM assistance. The authors take full responsibility for the work.

%% file: sec/X_suppl.tex
\clearpage
\setcounter{page}{1}
\maketitlesupplementary

\section{Event-only semantic segmentation on DSEC}
\label{sec:supp_seg}

Detection needs only a few boxes per image. Semantic segmentation \cite{fedorchenko_zehnder_2026eventsama} is
harder: it must assign a class to every pixel. It therefore tests
whether the compact SSM hidden states of \cref{sec:compact_states} keep
enough spatial detail for dense prediction. Each mixer reconstructs
features at all $N$ locations, so a standard dense decoder can be placed
on top of the same backbone.

\paragraph{Task and data.}
We use DSEC-Semantic~\cite{sun2022ess}. Each pixel belongs to one of 11
classes: background, building, fence, person, pole, road, sidewalk,
vegetation, car, wall, and traffic sign. Labels have a resolution of
$640\times440$, with 8,082 training and 2,809 test frames. The model
sees only events at inference.

\paragraph{Model.}
For each prediction, we accumulate events over 25, 100 and 300\,ms.
Short windows show fast motion, while long windows also show static
structure. Five temporal bins per window give 15 input channels. Our
compact SSM backbone turns this tensor into multiscale spatial features.
A causal temporal SSM combines the features of four successive event
tensors, so each prediction uses only current and past events. A
UPerNet decoder~\cite{xiao2018upernet} then predicts a class score for
every pixel.

\paragraph{Training.}
First, frozen DINOv3 image features~\cite{simeoni2025dinov3} teach an
event encoder to produce similar features from paired events. This uses
109,645 image--event samples from Cityscapes~\cite{cordts2016cityscapes},
KITTI~\cite{geiger2013kitti}, DECD~\cite{rebecq2019highspeed}, and
VisEvent/COESOT~\cite{wang2024visevent,tang2025coesot}. This encoder
initializes a semantic event teacher, which we pretrain together with
our compact SSM student for 31,250 updates. On DSEC, the student learns
from the labels and from the teacher's soft class predictions. The
distillation weight is 1 for the first 12,000 updates and 4 until
24,000. The final model averages the last three checkpoints and keeps
the batch-normalization statistics of the last one. Training uses 16
GH200 GPUs, BF16 precision, and a global batch of 32.

\paragraph{Selection without test data.}
We fit one additive score offset per class on the Zurich training
sequences \texttt{00\_a} and \texttt{08\_a} (1,726 frames). We confirm
the recipe on a second holdout, \texttt{01\_a} and \texttt{07\_a}
(2,144 frames). Each fold's teacher is trained without that fold. We
then retrain on all 8,082 training frames and evaluate each seed once
on the official test split. Seed~0 was chosen as the headline result
before evaluation. Each seed uses one native $640\times440$ view, with
no flipping or other test-time augmentation.

\paragraph{Metrics.}
For each class $c$, the intersection over union compares the pixels
predicted as $c$ with the pixels labelled as $c$:
\[
    \mathrm{IoU}_c=\frac{\mathrm{TP}_c}{\mathrm{TP}_c+\mathrm{FP}_c+\mathrm{FN}_c},
\]
where $\mathrm{TP}_c$, $\mathrm{FP}_c$ and $\mathrm{FN}_c$ count
correct, wrongly added, and missed pixels. The mean IoU (mIoU) averages
$\mathrm{IoU}_c$ over all 11 classes, so every class counts equally.
Pixel accuracy is the fraction of all labelled pixels whose class is
predicted correctly. Large classes such as road, building and
background dominate it, so a model can score high while missing thin
classes such as poles and signs. We accumulate counts over all 2,809
test frames before computing both metrics. We compare with other
methods only in mIoU, because it is the one metric that every paper
reports with the same definition. Accuracy scores in prior work do not
share one definition: ScaleEvent reports average class
accuracy~\cite{chen2026scaleevent}, and other papers do not always
state which accuracy they use. We therefore report our pixel accuracy
only for completeness.

\begin{table}[t]
\centering
\caption{\textbf{Our three seeds on the DSEC-Semantic test split.}
Mean $\pm$ population SD.}
\label{tab:supp_seg_seeds}
\begin{tabular}{@{}lcc@{}}
\toprule
Run & mIoU (\%) $\uparrow$ & Pixel acc.\ (\%) $\uparrow$ \\
\midrule
Seed 0 (prespecified) & 70.08 & 93.77 \\
Seed 1 & 69.89 & 93.78 \\
Seed 2 & 70.10 & 93.77 \\
\midrule
Mean $\pm$ SD & $\mathbf{70.02 \pm 0.10}$ & $93.77 \pm 0.01$ \\
\bottomrule
\end{tabular}
\end{table}

\begin{table}[t]
\centering
\caption{\textbf{Comparison on DSEC-Semantic.} Event-only methods,
mIoU (\%) as reported by each paper. Training data, supervision, and
test-time augmentation differ across papers.}
\label{tab:supp_seg_compare}
\begin{tabular}{@{}llc@{}}
\toprule
Method & Backbone & mIoU $\uparrow$ \\
\midrule
ESS (supervised)~\cite{sun2022ess} & E2VID & 53.29 \\
OpenESS~\cite{kong2024openess} & E2VID & 57.21 \\
ECDP~\cite{yang2023ecdp} & ResNet-50 & 59.16 \\
ECDDP~\cite{yang2024ecddp} & Swin-T & 61.25 \\
STP~\cite{liang2025stp} & Swin-T & 62.05 \\
TESPEC~\cite{mohammadi2025tespec} & recurrent Swin-T & 62.77 \\
GEP~\cite{cao2026gep} & ViT-B & 67.37 \\
Hyper-RED~\cite{wang2026hyperred} & ViT-L/16 & 69.04 \\
ScaleEvent~\cite{chen2026scaleevent} & ViT-L/16 & 69.65 \\
\midrule
Ours (3-seed mean) & compact SSM & \textbf{70.02} \\
\bottomrule
\end{tabular}
\end{table}

\paragraph{Results.}
\cref{tab:supp_seg_seeds} shows that the three seeds reach
$70.02\pm0.10$ mIoU, so the result is stable across seeds.
\cref{tab:supp_seg_compare} compares this with published results. Our
mean is 0.37 points above ScaleEvent-L, the strongest event-only method
we found, and 0.98 points above Hyper-RED-L. Both use a ViT-L/16
encoder distilled from DINOv3. The margin is small, and training data
and supervision differ across papers. We therefore read this result as
matching the strongest event-only methods, not as a clear improvement
over them. The class-level results show where the model differs. Fence
IoU averages 36.81\%, against 26.88\% for ScaleEvent-L. Wall IoU
averages 45.25\%, against 52.53\%.

%% file: main.bib
@inproceedings{wang2024crosskd,
  title={CrossKD: Cross-head knowledge distillation for object detection},
  author={Wang, Jiabao and Chen, Yuming and Zheng, Zhaohui and Li, Xiang and Cheng, Ming-Ming and Hou, Qibin},
  booktitle={2024 IEEE/CVF Conference on Computer Vision and Pattern Recognition (CVPR)},
  pages={16520--16530},
  year={2024},
  organization={IEEE}
}

@article{dao2024mamba2,
  title={Transformers are SSMs: Generalized models and efficient algorithms through structured state space duality},
  author={Dao, Tri and Gu, Albert},
  journal={Proceedings of the 41st International Conference on Machine Learning},
  year={2024}
}

@inproceedings{lee2025efficientvim,
  title={Efficientvim: Efficient vision mamba with hidden state mixer based state space duality},
  author={Lee, Sanghyeok and Choi, Joonmyung and Kim, Hyunwoo J},
  booktitle={2025 IEEE/CVF Conference on Computer Vision and Pattern Recognition (CVPR)},
  pages={14923--14933},
  year={2025},
  organization={IEEE}
}

@inproceedings{liang2025stp,
  title={Efficient event camera data pretraining with adaptive prompt fusion},
  author={Liang, Quanmin and Li, Qiang and Liu, Shuai and Cao, Xinzi and Lu, Jinyi and Yang, Feidiao and Zhang, Wei and Huang, Kai and Tian, Yonghong},
  booktitle={2025 IEEE/CVF International Conference on Computer Vision (ICCV)},
  pages={8656--8667},
  year={2025},
  organization={IEEE}
}

@inproceedings{fan2025eventpillars,
  title={Eventpillars: Pillar-based efficient representations for event data},
  author={Fan, Rui and Hao, Weidong and Guan, Juntao and Rui, Lai and Gu, Lin and Wu, Tong and Zeng, Fanhong and Zhu, Zhangming},
  booktitle={Proceedings of the AAAI Conference on Artificial Intelligence},
  volume={39},
  number={3},
  pages={2861--2869},
  year={2025}
}

@article{gehrig2024dagr,
  title={Low-latency automotive vision with event cameras},
  author={Gehrig, Daniel and Scaramuzza, Davide},
  journal={Nature},
  volume={629},
  number={8014},
  pages={1034--1040},
  year={2024},
  publisher={Nature Publishing Group UK, London}
}

@article{zhang2026efh,
  title={Event-Fused Hybrid ANN-SNN Architecture for Low-Latency Object Detection in Automotive Vision},
  author={Zhang, Chengjun and Zhang, Yuhao and Yu, Jisong and Yang, Jie and Sawan, Mohamad},
  journal={IEEE Robotics and Automation Letters},
  volume={11},
  number={3},
  pages={3622--3628},
  year={2026},
  publisher={IEEE}
}

@inproceedings{khitushkin2026spikeyolo,
  title={Using Spiking Neural Networks for Event and Multimodal Data Processing in Object Detection Tasks},
  author={Khitushkin, Kirill S and Isakov, Tim T and Bakhshiev, Aleksandr V},
  booktitle={International Conference on Neuroinformatics},
  pages={123--137},
  year={2025},
  organization={Springer}
}

@inproceedings{hao2026ssla,
  title={Low-latency event-based object detection with spatially-sparse linear attention},
  author={Hao, Haiqing and Sui, Zhipeng and Zou, Rong and Dai, Zijia and Zubi{\'c}, Nikola and Scaramuzza, Davide and Wang, Wenhui},
  booktitle={European Conference on Computer Vision},
  pages={20--38},
  year={2026},
  organization={Springer}
}

@inproceedings{gehrig2019est,
  title={End-to-end learning of representations for asynchronous event-based data},
  author={Gehrig, Daniel and Loquercio, Antonio and Derpanis, Konstantinos G and Scaramuzza, Davide},
  booktitle={Proceedings of the IEEE/CVF international conference on computer vision},
  pages={5633--5643},
  year={2019}
}

@article{perot2020red,
  title={Learning to detect objects with a 1 megapixel event camera},
  author={Perot, Etienne and De Tournemire, Pierre and Nitti, Davide and Masci, Jonathan and Sironi, Amos},
  journal={Advances in Neural Information Processing Systems},
  volume={33},
  pages={16639--16652},
  year={2020}
}

@inproceedings{gehrig2023rvt,
  title={Recurrent vision transformers for object detection with event cameras},
  author={Gehrig, Mathias and Scaramuzza, Davide},
  booktitle={2023 IEEE/CVF Conference on Computer Vision and Pattern Recognition (CVPR)},
  pages={13884--13893},
  year={2023},
  organization={IEEE}
}

@inproceedings{schaefer2022aegnn,
  title={Aegnn: Asynchronous event-based graph neural networks},
  author={Schaefer, Simon and Gehrig, Daniel and Scaramuzza, Davide},
  booktitle={2022 IEEE/CVF conference on computer vision and pattern recognition (CVPR)},
  pages={12361--12371},
  year={2022},
  organization={IEEE}
}

@inproceedings{li2025msd,
  title={Brain-inspired spiking neural networks for energy-efficient object detection},
  author={Li, Ziqi and Gao, Tao and An, Yisheng and Chen, Ting and Zhang, Jing and Wen, Yuanbo and Liu, Mengkun and Zhang, Qianxi},
  booktitle={2025 IEEE/CVF Conference on Computer Vision and Pattern Recognition (CVPR)},
  pages={3552--3562},
  year={2025},
  organization={IEEE}
}

@inproceedings{santambrogio2024farse,
  title={Farse-cnn: Fully asynchronous, recurrent and sparse event-based cnn},
  author={Santambrogio, Riccardo and Cannici, Marco and Matteucci, Matteo},
  booktitle={European conference on computer vision},
  pages={1--18},
  year={2024},
  organization={Springer}
}

@inproceedings{gu2024mamba,
  author={Gu, Albert and Dao, Tri},
  title={{Mamba}: Linear-Time Sequence Modeling with Selective State Spaces},
  booktitle={First Conference on Language Modeling}, year={2024},
}

@inproceedings{zhu2024visionmamba,
  title={Vision Mamba: Efficient Visual Representation Learning with Bidirectional State Space Model},
  author={Liang-Hui Zhu and Ben-Cheng Liao and Qian Zhang and Xin-Long Wang and Wen-Yu Liu and Xing-Gang Wang},
  booktitle={International Conference on Machine Learning},
  year={2024},
}

@inproceedings{yang2025smamba,
  title={Smamba: Sparse mamba for event-based object detection},
  author={Yang, Nan and Wang, Yang and Liu, Zhanwen and Li, Meng and An, Yisheng and Zhao, Xiangmo},
  booktitle={Proceedings of the AAAI conference on artificial intelligence},
  volume={39},
  number={9},
  pages={9229--9237},
  year={2025}
}

@article{yang2025focusmamba,
  title={Focus through motion: RGB-event collaborative token sparsification for efficient object detection},
  author={Yang, Nan and Wang, Yang and Liu, Zhanwen and Dai, Yuchao and Liu, Yang and Zhao, Xiangmo},
  journal={IEEE Transactions on Circuits and Systems for Video Technology},
  year={2026},
  publisher={IEEE}
}

@article{zhang2024faod,
  title={Frequency-adaptive low-latency object detection using events and frames},
  author={Zhang, Haitian and Wang, Xiangyuan and Xu, Chang and Wang, Xinya and Xu, Fang and Yu, Huai and Yu, Lei and Yang, Wen},
  journal={arXiv preprint arXiv:2412.04149},
  year={2024}
}

@inproceedings{li2025acgr,
  title={Asynchronous collaborative graph representation for frames and events},
  author={Li, Dianze and Li, Jianing and Liu, Xu and Fan, Xiaopeng and Tian, Yonghong},
  booktitle={2025 IEEE/CVF Conference on Computer Vision and Pattern Recognition (CVPR)},
  pages={1655--1666},
  year={2025},
  organization={IEEE}
}

@article{li2026hdiformer,
  title={Rethinking Neuromorphic Object Detection with Hybrid Dynamic Interaction Transformers: D. Li et al.},
  author={Li, Dianze and Li, Jianing and Liu, Xu and Zhou, Zhaokun and Fan, Xiaopeng and Tian, Yonghong},
  journal={International Journal of Computer Vision},
  volume={134},
  number={7},
  pages={340},
  year={2026},
  publisher={Springer}
}

@article{lu2025flexevent,
  title={Flexevent: towards flexible event-frame object detection at varying operational frequencies},
  author={Lu, Dongyue and Kong, Lingdong and Lee, Gim Hee and Simon Chane, Camille and Ooi, Wei},
  journal={Advances in Neural Information Processing Systems},
  volume={38},
  pages={89009--89044},
  year={2026}
}

@inproceedings{kim2021nimagenet,
  title={N-imagenet: Towards robust, fine-grained object recognition with event cameras},
  author={Kim, Junho and Bae, Jaehyeok and Park, Gangin and Zhang, Dongsu and Kim, Young Min},
  booktitle={2021 IEEE/CVF International Conference on Computer Vision (ICCV)},
  pages={2126--2136},
  year={2021},
  organization={IEEE}
}

@inproceedings{zubic2023ergo,
  title={From chaos comes order: Ordering event representations for object recognition and detection},
  author={Zubi{\'c}, Nikola and Gehrig, Daniel and Gehrig, Mathias and Scaramuzza, Davide},
  booktitle={2023 IEEE/CVF International Conference on Computer Vision (ICCV)},
  pages={12800--12810},
  year={2023},
  organization={IEEE}
}

@inproceedings{zubic2024eventssm,
  title={State space models for event cameras},
  author={Zubic, Nikola and Gehrig, Mathias and Scaramuzza, Davide},
  booktitle={Proceedings of the IEEE/CVF conference on computer vision and pattern recognition},
  pages={5819--5828},
  year={2024}
}

@inproceedings{rebecq2019e2vid,
  title={Events-to-video: Bringing modern computer vision to event cameras},
  author={Rebecq, Henri and Ranftl, Ren{\'e} and Koltun, Vladlen and Scaramuzza, Davide},
  booktitle={2019 IEEE/CVF Conference on Computer Vision and Pattern Recognition (CVPR)},
  pages={3852--3861},
  year={2019},
  organization={IEEE}
}

@inproceedings{gehrig2020videoevents,
  title={Video to events: Recycling video datasets for event cameras},
  author={Gehrig, Daniel and Gehrig, Mathias and Hidalgo-Carri{\'o}, Javier and Scaramuzza, Davide},
  booktitle={2020 IEEE/CVF Conference on Computer Vision and Pattern Recognition (CVPR)},
  pages={3583--3592},
  year={2020},
  organization={IEEE}
}

@inproceedings{yang2023ecdp,
  title={Event camera data pre-training},
  author={Yang, Yan and Pan, Liyuan and Liu, Liu},
  booktitle={2023 IEEE/CVF International Conference on Computer Vision (ICCV)},
  pages={10665--10675},
  year={2023},
  organization={IEEE}
}

@inproceedings{klenk2024mem,
  title={Masked event modeling: Self-supervised pretraining for event cameras},
  author={Klenk, Simon and Bonello, David and Koestler, Lukas and Araslanov, Nikita and Cremers, Daniel},
  booktitle={2024 IEEE/CVF Winter Conference on Applications of Computer Vision (WACV)},
  pages={2367--2377},
  year={2024},
  organization={IEEE}
}

@article{zhou2025pass,
  title={PASS: path-selective state space model for event-based recognition},
  author={Zhou, Jiazhou and Chen, Kanghao and Zhang, Lei and Wang, Lin},
  journal={Advances in Neural Information Processing Systems},
  volume={38},
  pages={168236--168264},
  year={2026}
}

@inproceedings{zhou2024eventbind,
  title={Eventbind: Learning a unified representation to bind them all for event-based open-world understanding},
  author={Zhou, Jiazhou and Zheng, Xu and Lyu, Yuanhuiyi and Wang, Lin},
  booktitle={European Conference on Computer Vision},
  pages={477--494},
  year={2024},
  organization={Springer}
}

@article{deTournemire2020gen1,
  title={A large scale event-based detection dataset for automotive},
  author={De Tournemire, Pierre and Nitti, Davide and Perot, Etienne and Migliore, Davide and Sironi, Amos},
  journal={arXiv preprint arXiv:2001.08499},
  year={2020}
}

@article{orchard2015ncaltech,
  title={Converting static image datasets to spiking neuromorphic datasets using saccades},
  author={Orchard, Garrick and Jayawant, Ajinkya and Cohen, Gregory K and Thakor, Nitish},
  journal={Frontiers in neuroscience},
  volume={9},
  pages={437},
  year={2015},
  publisher={Frontiers Media SA}
}

@inproceedings{iacono2018inceptionssd,
  title={Towards event-driven object detection with off-the-shelf deep learning},
  author={Iacono, Massimiliano and Weber, Stefan and Glover, Arren and Bartolozzi, Chiara},
  booktitle={2018 IEEE/RSJ International Conference on Intelligent Robots and Systems (IROS)},
  pages={1--9},
  year={2018},
  organization={IEEE}
}

@inproceedings{chen2018eventsrrc,
  title={Pseudo-labels for supervised learning on dynamic vision sensor data, applied to object detection under ego-motion},
  author={Chen, Nicholas FY},
  booktitle={2018 IEEE/CVF Conference on Computer Vision and Pattern Recognition Workshops (CVPRW)},
  pages={757--75709},
  year={2018},
  organization={IEEE}
}

@inproceedings{cannici2020matrixlstm,
  title={A differentiable recurrent surface for asynchronous event-based data},
  author={Cannici, Marco and Ciccone, Marco and Romanoni, Andrea and Matteucci, Matteo},
  booktitle={European Conference on Computer Vision},
  pages={136--152},
  year={2020},
  organization={Springer}
}

@inproceedings{jiang2019eventsyolo,
  title={Mixed frame-/event-driven fast pedestrian detection},
  author={Jiang, Zhuangyi and Xia, Pengfei and Huang, Kai and Stechele, Walter and Chen, Guang and Bing, Zhenshan and Knoll, Alois},
  booktitle={2019 International Conference on Robotics and Automation (ICRA)},
  pages={8332--8338},
  year={2019},
  organization={IEEE}
}

@inproceedings{li2021nvss,
  title={Graph-based asynchronous event processing for rapid object recognition},
  author={Li, Yijin and Zhou, Han and Yang, Bangbang and Zhang, Ye and Cui, Zhaopeng and Bao, Hujun and Zhang, Guofeng},
  booktitle={2021 IEEE/CVF International Conference on Computer Vision (ICCV)},
  pages={914--923},
  year={2021},
  organization={IEEE}
}

@inproceedings{messikommer2020asynet,
  title={Event-based asynchronous sparse convolutional networks},
  author={Messikommer, Nico and Gehrig, Daniel and Loquercio, Antonio and Scaramuzza, Davide},
  booktitle={European Conference on Computer Vision},
  pages={415--431},
  year={2020},
  organization={Springer}
}

@inproceedings{hao2026eva,
  title={Maximizing asynchronicity in event-based neural networks},
  author={Hao, Haiqing and Zubic, Nikola and He, Weihua and Sui, Zhipeng and Scaramuzza, Davide and Wang, Wenhui},
  booktitle={International Conference on Learning Representations},
  volume={2026},
  pages={123834--123855},
  year={2026}
}

@inproceedings{peng2023get,
  title={Get: Group event transformer for event-based vision},
  author={Peng, Yansong and Zhang, Yueyi and Xiong, Zhiwei and Sun, Xiaoyan and Wu, Feng},
  booktitle={2023 IEEE/CVF International Conference on Computer Vision (ICCV)},
  pages={6015--6025},
  year={2023},
  organization={IEEE}
}

@inproceedings{peng2024sast,
  title={Scene adaptive sparse transformer for event-based object detection},
  author={Peng, Yansong and Li, Hebei and Zhang, Yueyi and Sun, Xiaoyan and Wu, Feng},
  booktitle={2024 IEEE/CVF Conference on Computer Vision and Pattern Recognition (CVPR)},
  pages={16794--16804},
  year={2024},
  organization={IEEE}
}

@article{chen2025ehgcn,
  title={Ehgcn: Hierarchical euclidean-hyperbolic fusion via motion-aware gcn for hybrid event stream perception},
  author={Chen, Haosheng and Luo, Lian and Mo, Mengjingcheng and Wu, Zhanjie and Gan, Ji and Leng, Jiaxu and Gao, Xinbo},
  journal={arXiv preprint arXiv:2504.16616},
  year={2025}
}

@article{gehrig2021dsec,
  title={Dsec: A stereo event camera dataset for driving scenarios},
  author={Gehrig, Mathias and Aarents, Willem and Gehrig, Daniel and Scaramuzza, Davide},
  journal={IEEE Robotics and Automation Letters},
  volume={6},
  number={3},
  pages={4947--4954},
  year={2021},
  publisher={IEEE}
}

@inproceedings{zhang2021varifocalnet,
  title={Varifocalnet: An iou-aware dense object detector},
  author={Zhang, Haoyang and Wang, Ying and Dayoub, Feras and Sunderhauf, Niko},
  booktitle={Proceedings of the IEEE/CVF conference on computer vision and pattern recognition},
  pages={8514--8523},
  year={2021}
}

@article{rezatofighi2019giou,
  title={Generalized Intersection Over Union: A Metric and a Loss for Bounding Box Regression},
  author={Hamid Rezatofighi and Nathan Tsoi and JunYoung Gwak and Amir Sadeghian and Ian D. Reid and Silvio Savarese},
  journal={2019 IEEE/CVF Conference on Computer Vision and Pattern Recognition (CVPR)},
  year={2019},
  pages={658-666},
}

@article{li2020gfl,
  title={Generalized focal loss: Learning qualified and distributed bounding boxes for dense object detection},
  author={Li, Xiang and Wang, Wenhai and Wu, Lijun and Chen, Shuo and Hu, Xiaolin and Li, Jun and Tang, Jinhui and Yang, Jian},
  journal={Advances in neural information processing systems},
  volume={33},
  pages={21002--21012},
  year={2020}
}

@inproceedings{lin2017retinanet,
  title={Focal loss for dense object detection},
  author={Lin, Tsung-Yi and Goyal, Priya and Girshick, Ross and He, Kaiming and Doll{\'a}r, Piotr},
  booktitle={Proceedings of the IEEE international conference on computer vision},
  pages={2980--2988},
  year={2017}
}

@inproceedings{loshchilov2019adamw,
  title={Decoupled Weight Decay Regularization},
  author={Ilya Loshchilov and Frank Hutter},
  booktitle={International Conference on Learning Representations},
  year={2017},
}

@article{soydan2024s7,
  title={S7: Selective and simplified state space layers for sequence modeling},
  author={Soydan, Taylan and Zubi{\'c}, Nikola and Messikommer, Nico and Mishra, Siddhartha and Scaramuzza, Davide},
  journal={arXiv preprint arXiv:2410.03464},
  year={2024}
}

@inproceedings{zubic2025gg,
  title={Gg-ssms: Graph-generating state space models},
  author={Zubic, Nikola and Scaramuzza, Davide},
  booktitle={Proceedings of the IEEE/CVF conference on computer vision and pattern recognition},
  pages={28863--28873},
  year={2025}
}

@inproceedings{raju2025perturbed,
  title={Perturbed state space feature encoders for optical flow with event cameras},
  author={Raju, Gokul Raju Govinda and Zubic, Nikola and Cannici, Marco and Scaramuzza, Davide},
  booktitle={2025 IEEE/CVF Conference on Computer Vision and Pattern Recognition Workshops (CVPRW)},
  pages={5072--5081},
  year={2025},
  organization={IEEE}
}

@article{zubic2026regularity,
  title={Regularity and Stability Properties of Selective SSMs with Discontinuous Gating},
  author={Zubi{\'c}, Nikola and Scaramuzza, Davide},
  journal={Transactions on Machine Learning Research},
  year={2026}
}

@inproceedings{lu2026hybrid,
  title={Hybrid Event--Frame Sensors: Modeling, Calibration, and Simulation},
  author={Lu, Yunfan and Messikommer, Nico and Xu, Xiaogang and Chen, Liming and Chen, Yuhan and Zubi{\'c}, Nikola and Scaramuzza, Davide and Xiong, Hui},
  booktitle={European Conference on Computer Vision},
  pages={167--187},
  year={2026},
  organization={Springer}
}

@article{zubic2026expressive,
  title={On the expressive power and limitations of multi-layer ssms},
  author={Zubi{\'c}, Nikola and Li, Qian and Wang, Yuyi and Scaramuzza, Davide},
  journal={Advances in Neural Information Processing Systems},
  year={2026}
}

@inproceedings{fedorchenko_zehnder_2026eventsama,   
author={Fedorchenko, Sergei and Zehnder, Max and Zubi{\'c}, Nikola and Zou, Rong and Scaramuzza, Davide},   
title={{Event-SAMA}: Language-Grounded Video Object Segmentation on Neuromorphic Event Streams},
booktitle={Conference on Neural Information Processing Systems Workshops ({NeurIPSW})},
year={2026}
}

@inproceedings{sun2022ess,
  title={Ess: Learning event-based semantic segmentation from still images},
  author={Sun, Zhaoning and Messikommer, Nico and Gehrig, Daniel and Scaramuzza, Davide},
  booktitle={European Conference on Computer Vision},
  pages={341--357},
  year={2022},
  organization={Springer}
}

@article{chen2026scaleevent,
  title={Scaling Dense Event-Stream Pretraining from Visual Foundation Models},
  author={Chen, Zhiwen and Hou, Junhui and Zhu, Zhiyu and Wu, Jinjian and Shi, Guangming},
  journal={arXiv preprint arXiv:2603.03969},
  year={2026}
}

@article{wang2026hyperred,
  title={Hyper-RED: Scalable Event Pre-training via Semantic Hypergraph Distillation},
  author={Wang, Meisen and Tian, Zhiqiang and Bao, Wei and Wang, Chengjie and Du, Shaoyi and Li, Siqi},
  journal={arXiv preprint arXiv:2609.16811},
  year={2026}
}

@article{cao2026gep,
  title={Generative Event Pretraining with Foundation Model Alignment},
  author={Cao, Jianwen and Xing, Jiaxu and Messikommer, Nico and Scaramuzza, Davide},
  journal={arXiv preprint arXiv:2603.23032},
  year={2026}
}

@inproceedings{mohammadi2025tespec,
  title={Tespec: Temporally-enhanced self-supervised pretraining for event cameras},
  author={Mohammadi, Mohammad and Wu, Ziyi and Gilitschenski, Igor},
  booktitle={2025 IEEE/CVF International Conference on Computer Vision (ICCV)},
  pages={7782--7793},
  year={2025},
  organization={IEEE}
}

@inproceedings{kong2024openess,
  title={Openess: Event-based semantic scene understanding with open vocabularies},
  author={Kong, Lingdong and Liu, Youquan and Ng, Lai Xing and Cottereau, Benoit R and Ooi, Wei Tsang},
  booktitle={2024 IEEE/CVF Conference on Computer Vision and Pattern Recognition (CVPR)},
  pages={15686--15698},
  year={2024},
  organization={IEEE}
}

@inproceedings{xiao2018upernet,
  title={Unified perceptual parsing for scene understanding},
  author={Xiao, Tete and Liu, Yingcheng and Zhou, Bolei and Jiang, Yuning and Sun, Jian},
  booktitle={European conference on computer vision},
  pages={432--448},
  year={2018},
  organization={Springer}
}

@article{simeoni2025dinov3,
  title={Dinov3},
  author={Sim{\'e}oni, Oriane and Vo, Huy V and Seitzer, Maximilian and Baldassarre, Federico and Oquab, Maxime and Jose, Cijo and Khalidov, Vasil and Szafraniec, Marc and Yi, Seungeun and Ramamonjisoa, Micha{\"e}l and others},
  journal={arXiv preprint arXiv:2508.10104},
  year={2025}
}

@inproceedings{cordts2016cityscapes,
  title={The cityscapes dataset for semantic urban scene understanding},
  author={Cordts, Marius and Omran, Mohamed and Ramos, Sebastian and Rehfeld, Timo and Enzweiler, Markus and Benenson, Rodrigo and Franke, Uwe and Roth, Stefan and Schiele, Bernt},
  booktitle={Proceedings of the IEEE conference on computer vision and pattern recognition},
  pages={3213--3223},
  year={2016}
}

@article{geiger2013kitti,
  title={Vision meets robotics: The kitti dataset},
  author={Geiger, Andreas and Lenz, Philip and Stiller, Christoph and Urtasun, Raquel},
  journal={The international journal of robotics research},
  volume={32},
  number={11},
  pages={1231--1237},
  year={2013},
  publisher={Sage Publications, Sage UK: London, England}
}

@article{tang2025coesot,
  title={Revisiting color-event based tracking: A unified network, dataset, and metric},
  author={Tang, Chuanming and Wang, Xiao and Huang, Ju and Jiang, Bo and Zhu, Lin and Chen, Shifeng and Zhang, Jianlin and Wang, Yaowei and Tian, Yonghong},
  journal={Pattern Recognition},
  pages={112718},
  year={2025},
  publisher={Elsevier}
}

@article{rebecq2019highspeed,
  title={High speed and high dynamic range video with an event camera},
  author={Rebecq, Henri and Ranftl, Ren{\'e} and Koltun, Vladlen and Scaramuzza, Davide},
  journal={IEEE transactions on pattern analysis and machine intelligence},
  volume={43},
  number={6},
  pages={1964--1980},
  year={2019},
  publisher={IEEE}
}

@article{wang2024visevent,
  author  = {Wang, Xiao and Li, Jianing and Zhu, Lin and Zhang, Zhipeng and Chen, Zhe and Li, Xin and Wang, Yaowei and Tian, Yonghong and Wu, Feng},
  title   = {{VisEvent}: Reliable Object Tracking via Collaboration of Frame and Event Flows},
  journal = {IEEE Transactions on Cybernetics},
  year    = {2023}
}

@inproceedings{yang2024ecddp,
  title={Event camera data dense pre-training},
  author={Yang, Yan and Pan, Liyuan and Liu, Liu},
  booktitle={European Conference on Computer Vision},
  pages={292--310},
  year={2024},
  organization={Springer}
}
